\documentclass{article}

\usepackage{microtype}
\usepackage{graphicx}
\usepackage{subfigure}
\usepackage{booktabs} % for professional tables
\usepackage{amsmath}
\usepackage{amssymb}
\usepackage{amsfonts}
\usepackage{float}

\usepackage[noend]{algpseudocode}

\DeclareRobustCommand{\uvec}[1]{{%
  \ifcsname uvec#1\endcsname
     \csname uvec#1\endcsname
   \else
    \bm{\hat{\mathbf{#1}}}%
   \fi
}}

\makeatletter
\def\BState{\State\hskip-\ALG@thistlm}
\makeatother

\usepackage[accepted]{icml2018}
\begin{document}

\twocolumn[
\icmltitle{Modeling Others using Oneself in Multi-Agent Reinforcement Learning}

% \icmltitle{Inferring the Intentions of Other Agents in \\
%           Multi-Agent Reinforcement Learning}

% \icmltitle{Inferring Other Agents' Goals in \\
%           Multi-Agent Reinforcement Learning}




\begin{icmlauthorlist}
\icmlauthor{Roberta Raileanu}{nyu}
\icmlauthor{Emily Denton}{nyu}
\icmlauthor{Arthur Szlam}{fair}
\icmlauthor{Rob Fergus}{nyu,fair}
\end{icmlauthorlist}

\icmlaffiliation{nyu}{Department of Computer Science, New York University, New York City, New York, USA}
\icmlaffiliation{fair}{Facebook AI Research, Facebook Inc., New York City, New York, USA}

\icmlcorrespondingauthor{Roberta Raileanu}{raileanu@cs.nyu.edu}
% \icmlcorrespondingauthor{rr}{raileanu@cs.nyu.edu}
% \icmlcorrespondingauthor{ed}{denton@cs.nyu.edu}
% \icmlcorrespondingauthor{as}{aszlam@fb.co}
% \icmlcorrespondingauthor{rf}{robfergus@fb.com}


\vskip 0.3in
]



\printAffiliationsAndNotice{}  % leave blank if no need to mention equal contribution

%%%%%%%%%%%%%%%%%%%%%%%%%%% ABSTRACT %%%%%%%%%%%%%%%%%%%%%%%%%%%%%%%%%%

\begin{abstract}
We consider the multi-agent reinforcement learning setting with imperfect information in which 
each agent is trying to maximize its own utility. The reward function depends on the hidden state (or goal) of both agents, 
so the agents must infer the other players' hidden goals from their observed behavior in order to solve the tasks. 
We propose a new approach for learning in these domains: Self Other-Modeling (\textsc{SOM}), 
in which an agent uses its own policy to predict the other agent's actions and update its belief of their hidden state in an online manner. 
We evaluate this approach on three different tasks and show that the agents are able to learn better policies 
using their estimate of the other players' hidden states, in both cooperative and adversarial settings. 

% XXX: should I add a sentence related to the baselines compared against? 
    
\end{abstract}

% TODO INTRO: add more motivation for the problem and for the model 
% ... examples of realistic scenarios in which this is important to have and useful to use your self to model others
%%%%%%%%%%%%%%%%%%%%%%%%%%% INTRODUCTION %%%%%%%%%%%%%%%%%%%%%%%%%%%%%%%%%%
\section{Introduction}
\label{introduction}

Reasoning about other agents' intentions and being able to 
predict their behavior is important in multi-agent systems, 
in which the agents might have a diverse, and sometimes competing, set of goals.
This remains a challenging problem due to the inherent non-stationarity of such domains.

In this paper, we introduce a new approach for estimating the other agents' unknown goals from their behavior 
and using those estimates to choose actions. 
%We demonstrate that the agents are able to estimate the other agents' hidden states, 
%which enables them to converge to better policies and gain higher reward. 
We demonstrate that in the proposed tasks, using an explicit model of the other player in the game 
leads to better performance than simply considering the other agent to be part of the environment. 

We frame the problem as a (not-necessarily zero-sum) two-player stochastic game \cite{Shapley1095}, otherwise known as a two-player Markov game, 
in which the agents have full visibility of the environment, 
but no explicit knowledge about other agents' goals and there is no communication channel. 
The reward received by each agent at the end of an episode depends on the goals of both agents,
so the optimal policy of each agent must take into account both of their goals.

% The reward structure of the games depends on both agents'
% goals, so it is essential for the agents to infer the other player's hidden aims
% as early as possible during each episode in order to maximize their rewards. 

% The key idea of this work is that as a first approximation, to understand what a fellow player in the game is doing, an agent should ask itself ``what would be my goal if I had acted as my fellow had?".  We instantiate this idea by parametrizing the agent's action and value functions with a (multi-layer recurrent) neural network that takes the state and a goal as an input.  As the agent plays the game, it infers its fellow agent's unknown goal by directly optimizing over the goal (using its own action function) to maximize the likelihood of the fellow's actions.  
%\rob{inconsistency here: we use "other" in rest of paper. can we use that here too?}
%\roberta{changed fellow to other}

Research in cognitive science suggests that humans maintain models of other people they interact with,
which capture their goals, beliefs, or preferences \cite{gopnik1992child, premack1978does}. 
In some cases, humans use their own mental process to simulate others' behavior by adopting their perspective
\cite{gordon1986folk, gallese1998mirror}. This allows them to understand others' intentions or motives and act accordingly in social settings. 
Inspired by these studies, the key idea of our approach is that as a first approximation, to understand what the other player in the game is doing, an agent should ask itself ``what would be my goal if I had acted as the other player had?". We instantiate this idea by parametrizing the agent's action and value functions with a (multi-layer recurrent) neural network that takes the state and a goal as an input. As the agent plays the game, it infers the other agent's unknown goal by directly optimizing over the goal (using its own action function) to maximize the likelihood of the other's actions.

\section{Approach}%{}Modeling Other Agents Using Oneself}
\label{approach}

{\bf Background}: A Markov game for two agents is defined by a set of states
$\mathcal{S}$ describing the possible configurations of all agents, 
a set of actions $\mathcal{A}_1, \mathcal{A}_2$ and a set of observations
$\mathcal{O}_1, \mathcal{O}_2$ for each agents, and a transition function
$\mathcal{T}: \mathcal {S}\times \mathcal{A}_1 \times \mathcal{A}_2 \rightarrow \mathcal{S}$
which gives the probability distribution
on the next state as a function of current state and actions. 
Each agent i chooses actions by sampling from a stochastic policy
$\pi_{\theta_i}: \mathcal{S} \times \mathcal{A}_i \rightarrow [0,1]$. 
Each agent has a reward function 
which depends on agent's state and action: $r_i: \mathcal{S} \times \mathcal{A}_i \rightarrow \mathbb{R}$. 
Each agent i tries to maximize its own total expected return $R_i = \sum_{t = 0}^T \gamma^t r_i^t$, 
where $\gamma$ is a discount factor and T is the time horizon. 
In this work, we consider both cooperative, as well as adversarial settings. 

We now describe Self Other-Modeling (SOM), a new approach for inferring 
the other agents' goals in an online fashion during an episode 
and using these estimates to choose actions. 
To decide an action and to estimate the value of a state, 
we use a neural network $f$ that takes as input its own goal $z_{self}$, 
an estimate of the other player's goal $\tilde{z}_{other}$, and the observation state from its own perspective $s_{self}$,
and outputs a probability distribution over actions $\pi$ and a value estimate $V$, such that
for each agent i playing the game we have:

%\begin{equation*} \left(\pi,V\right) = f(s, z_{self}, \tilde{z}_{other}; \; \theta) \; .
%\end{equation*}
%\rob{as discussed, we need to make clear that all this is for a single agent. different agents will have different $\theta$. either we say it in text, or we add $^i$.}%\roberta{better?} 
\begin{equation*} \begin{bmatrix}\pi^i\\V^i\end{bmatrix} = f^i(s_{self}^i, z_{self}^i, \tilde{z}_{other}^i; \; \theta^i) \; .
\end{equation*}

Here $\theta^i$ are agent $i$'s parameters for $f$, which has one softmax output for the policy, 
one linear output for the value function, and all the non-output layers shared. 
The actions are sampled from the policy $\pi$. The observation state $s_{self}^i$ explicitly contains
the location of the acting agent (the one whose action is decided by $f^i$), as well as the location of the other agent.

Because an agent computes both its own actions and values, as well as estimates of the other agent's, 
each agent has  two networks (omitting the agent index $i$ for brevity):
\begin{equation}\label{eq:self} f_{self}(s_{self}, z_{self}, \tilde{z}_{other}; \; \theta_{self})\end{equation} and 
\begin{equation}\label{eq:other} f_{other}(s_{other}, \tilde{z}_{other}, z_{self}; \; \theta_{self}) \;.\end{equation}
The two networks are used in different ways: $f_{self}$ is used for computing the agent's own actions and values, and operates in a feed-forward manner. The agent uses $f_{other}$  to infer the other agent's goal via an optimization over $\tilde{z}_{other}$ given the other agent's observed actions.

%\emily{it seems a bit confusing to start out by describing each agent as having two networks..I think maybe we should start out describing how an agent uses its own policy to make inferences about the other agent, and *then * explain how the inference usage of the policy network is different from acting with the policy network. Also I think here is a good time to explain in words that $f_{other}$ is never updated using the opponents actions (or in other words does not learn during inference) but rather is used to optimize over an input vector}

We propose that each agent models the behavior of the other player using its own policy,
so that the parameters of $f_{other}$ are the same as the parameters of $f_{self}$.
However, note that the two networks differ in their relative placement of the inputs $z_{self}$ and $\tilde{z}_{other}$.
Additionally, since the environment is fully observed, the observation state of the two agents differs only by the specification of the agent's identity on the map (i.e. each agent will be able to distinguish between its own location and the other's location). Hence, in acting mode, the network $f_{self}$ will take as input $s_{self}$ and in inference mode, the network  $f_{other}$ will take as input $s_{other}$.

% work in progress...

\begin{algorithm}
\caption{SOM training for one episode}
\label{alg:som}
\begin{algorithmic}[1]
\Procedure{Self Other-Modeling}{}
    % initializations for each episode
    % initialize the estimate of the other agent's goal - for each player
    % \vspace{-0.5em}   
    \For {k := 1, num\_players}
        \State $\textcolor{red}{\tilde{z}_{other}^k}\gets \frac{1}{ngoals}\textbf{1}_{ngoals}$
        % \State $\textcolor{red}{\tilde{z}_{other}^i} \sim U(0, ngoals)$
    \EndFor
    % \vspace{-0.5em}   
    \State game.reset()
    % \vspace{-0.5em}   
    % episode loop
    \For{step := 1, episode\_length}
            \State $i \gets game.get\_acting\_agent()$
            \State $j \gets game.get\_non\_acting\_agent()$% \State \rob{yes, you need to explain}
            \State $s_{self}^i \gets game.get\_state()$
            \State $s_{other}^j \gets game.get\_state()$
            % \State $\textcolor{red}{\tilde{z}_{other}^{OH, i}} = argmax(\textcolor{red}{\tilde{z}_{other}^i}) \; \uvec{i}$
            \State $\textcolor{red}{\tilde{z}_{other}^{OH, i}} = one\_hot[argmax(\textcolor{red}{\tilde{z}_{other}^i})]$
            \State $\pi_{self}^i, V_{self}^i \hspace*{-0.1mm} \gets f_{self}^i(s_{self}^i, z_{self}^i, \textcolor{red}{\tilde{z}_{other}^{OH, i}}; \theta_{self}^i)$
            \State $a_{self}^i \sim \pi_{self}^i$
            \State $game.action(a_{self}^i)$
                % \vspace{-0.5em}   
                % inference loop
                \For{k : = 1, num\_inference\_steps}
                    \State $\textcolor{red}{\tilde{z}_{other}^{G, j}} = gumbel\_softmax(\textcolor{red}{\tilde{z}_{other}^j})$
                    \State $\tilde{\pi}_{other}^j \hspace*{-1.2mm} \gets f_{other}^j(s_{other}^j, \textcolor{red}{\tilde{z}_{other}^{G, j}}, {z}_{self}^j; \theta_{self}^j)$ 
                    \State $loss = cross\_entropy\_loss(\tilde{\pi}_{other}^j, \, a_{self}^i)$
                    \State $loss.backward()$
                    \State $update(\textcolor{red}{\tilde{z}_{other}^j})$
                % end of inference loop
                % \vspace{-0.5em}   
                \EndFor
    % \vspace{-0.5em}   
    % end of episode
   \EndFor
    % update the parameters of both models at the end of the episode
    \For {k := 1, num\_players}
        \State $policy.update(\theta_{self}^k)$
    \EndFor
\EndProcedure
\end{algorithmic}
\end{algorithm}

% so that $\tilde{\theta}_{other} = \theta_{self}$ at all times, from the perspective of each agent.

% XXX: is this too speculative, the claims are not supported by results in this paper 

% TODO: argue why using your self model is a good idea and in which situations it would be useful / more useful than having a separate model

\begin{figure}[htbp]
	    \centerline{%
		    \includegraphics[width=0.7\linewidth]{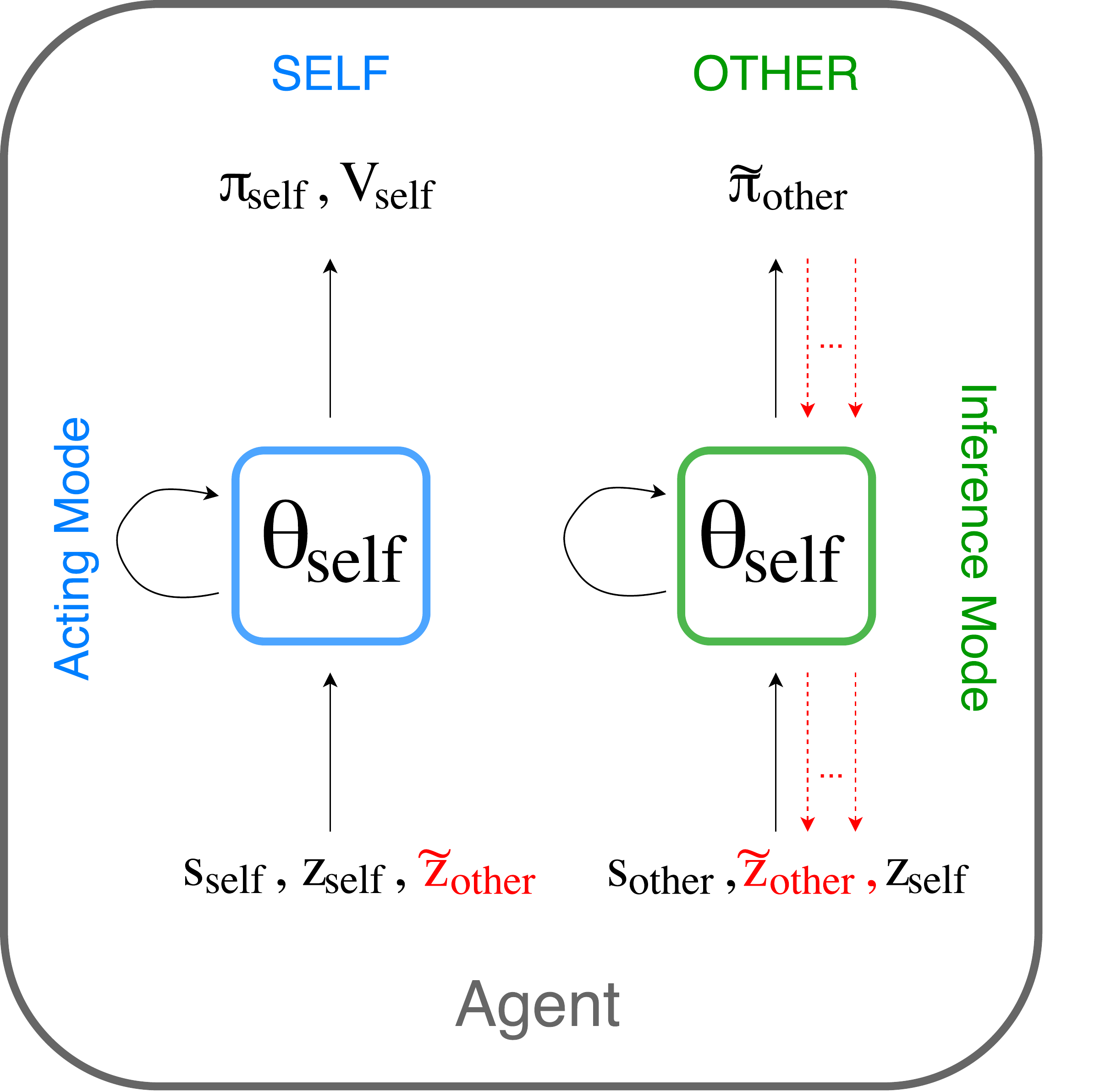}%
		}%
		\caption{Our Self Other-Model (SOM) architecture for a given agent.}
		\label{fig:model}
\end{figure}

At each step of the game, the agent needs to infer $\tilde{z}_{other}$ in order to input its estimate into
\eqref{eq:self} and choose its action. 
For this purpose, at each step, the agent observes the other taking an action and, at the next step, 
the agent uses the previously observed action of the other as supervision,
in order to back-propagate through \eqref{eq:other} and optimize over $\tilde{z}_{other}$. 
Figure~\ref{fig:model} illustrates this technique. 

% \rob{Refer to Figure 1 somewhere here}.
% For this purpose, the agent optimizes over 
% $\tilde{z}_{other}$ by back-propagating through \eqref{eq:other}, using supervision of the observed action the other agent took at the previous step. 

% \rob{something about this being visible}
The number of steps taken by the optimizer in this inference procedure is 
a hyperparameter that can be varied depending on the game. 
Hence, the estimate of the other agent's goal $\tilde{z}_{other}$ is updated multiple times at each step during the game.  The parameters ${\theta}_{self}$ are 
updated at the end of each episode
using Asynchronous Advantage Actor-Critic (A3C) \cite{mnih2016asynchronous} with reward signal obtained by the self agent.

% \rob{This paragraph is clear but you should refer to Algorithm 1 and the symbols used here}
%\roberta{is this more clear?}
Algorithm~\ref{alg:som} represents the pseudo-code for training SOM agents for one episode.
Since the goals are discrete in all the tasks considered here, 
the agent's goal $\tilde{z}_{self}$ is encoded as a one-hot vector of dimension equal to the total number of possible goals in the game. 
% \rob{assume discrete goals}
The embedding of the other player's goal $\tilde{z}_{other}$ has the same dimension.
In order to estimate the gradients going through $\tilde{z}_{other}$, which is a discrete variable and thus non-differentiable, 
we replace it with a differentiable sample from the Gumbel-Softmax distribution \cite{jang2016categorical, maddison2016concrete},
$\tilde{z}_{other}^{G}$. 
This reparametrization trick was shown to efficiently produce low-variance biased gradients. 
After optimizing $\tilde{z}_{other}$ at each step using this method, $\tilde{z}_{other}$ usually deviates
from a one-hot vector. At the next step, $f_{self}$ takes as input the one-hot vector $\tilde{z}_{other}^{OH}$ corresponding to the \textit{argmax} of the previously updated $\tilde{z}_{other}$. 

% \rob{some of this stuff, particuarly the architechture details should be earlier at end of approach section. we should certainly discuss recurrence in the model and this belong in the approach, for sure. I think maybe we should also have figure?}
The agents' policies are parametrized by long short-term memory (LSTM) cells \cite{hochreiter1997long} with 
two fully-connected linear layers,
and exponential linear unit (ELU) \cite{clevert2015fast} activations. 
The weights of the networks are initialized with semi-orthogonal matrices,
as described in \cite{saxe2013exact} and zero bias.

% \emily{This is my fix of the paragraph in question below.. feel free to remove / change if you dont like it. }
% Due to the recurrence of $f_{other}$, special care must be taken when the number of inference steps is $> 1$. Under this setting, we save the recurrent state of $f_{other}$ obtained by the first forward pass, and re-initialize the state to this value for every subsequent inference step. This procedure ensures $f_{other}$ is unrolled the same number of steps during both acting and inference mode.\roberta{this is much better! thanks!!}

% \roberta{using Emily's paragraph w. small modifications}
Due to the recurrence of $f_{other}$, special care must be taken when the number of inference steps is $> 1$. Under this setting, at each step in the game, we save the recurrent state of $f_{other}$ before the first forward pass in inference mode, and initialize the recurrent state to this value for every inference step. This procedure ensures $f_{other}$ is unrolled the same number of steps during both acting and inference mode.

\section{Related Work}
\label{relatedwork}

Opponent modeling has been extensively studied in 
games of imperfect information. However, most previous approaches focus on
developing models with domain-specific probabilistic priors 
or strategy parametrizations.
In contrast, our work proposes a more general framework for opponent modeling.
\cite{davidson1999using} uses an MLP to predict
opponent actions given a game history, but the agents
cannot adapt to their opponents' behavior in an online manner.
\cite{lockett2007evolving} designs a neural network
architecture to identify the opponent type
by learning a mixture of weights over a 
given set of cardinal opponents. However, the 
game does not unfold within the reinforcement learning framework.

A large body of work in deep multi-agent RL focuses on partially visible,
fully cooperative settings \cite{foerster2016learningA, foerster2016learningB, omidshafiei2017deep}
and emergent communication \cite{ lazaridou2016multi, foerster2016learningA, sukhbaatar2016learning, das2017learning, mordatch2017emergence}
Our setting is different since we do not allow any communication among the agents, so the players have to indirectly
reason about their opponents' intentions from their observed behavior.
In contrast, \cite{leibo2017multi} considers semi-cooperative multi-agent environments in which the agents develop
cooperative and competitive strategies depending on the task type and reward structure. 
Similarly, \cite{lowe2017multi} proposes a centralized actor-critic architecture for efficient training in 
settings with such mixed strategies. 
\cite{lerer2017maintaining} design RL agents that are able to maintain cooperation in 
complex social dilemmas by generalizing a well-known game theoretic strategy called tit-for-tat \cite{axelrod2006agent}
to multi-agent Markov games. Recent work in cognitive science attempts to understand human decision-making by using a hierarchical model of social agency that infers the intentions of other human agents in order to decide whether to play a cooperative or competitive strategy \cite{kleiman2016coordinate}. However, none of these papers design algorithms that explicitly
model other artificial agents in the environment or estimate their intentions, with the purpose of improve their decision making. 

The field of inverse reinforcement learning (IRL) \cite{russell1998learning, ng2000algorithms, abbeel2004apprenticeship},
is also related to the problem considered here. IRL's aim is to infer the reward function of an agent by observing its behavior,
which is assumed to be nearly optimal. In contrast, our approach uses the observed actions of the other player to directly infer
its goal in an online manner, which is then used by the agent when acting in the environment. 
This avoids the need for collecting offline samples of the other's (state, action) pairs 
in order to estimate its reward function and then use this to learn a separate policy that maximizes that utility. 
The more recent papers by \cite{hadfield2016cooperative, hadfield2017inverse}
are also concerned with the problem of inferring others' intentions, but their focus is on human-robot interaction and value alignment.
Motivated by similar goals, \cite{chandrasekaran2017takes} consider the problem of building a theory of AI's mind, in order to improve 
human-AI interaction and the interpretability of AI systems. For this purpose, they show that people can be trained
to predict the responses of a Visual Question Answering model, using a small number of examples. 

The closest work to ours is \cite{foerster2017learning} and \cite{he2016opponent}. 
\cite{foerster2017learning} designs RL agents that take into account the learning of
other agents in the environment when updating their own policies. This enables the agents to
discover self-interested yet collaborative strategies such as tit-for-that in the iterated prisoner's dilemma.
While our work does not explicitly attempt to shape the learning of other agents, it has the advantage that
the agents can update their beliefs during an episode and change their strategies in an online manner
to gain more reward. Our setting is also different in that it considers that each agent has some
hidden information needed by their the other player in order to maximize its return. 

Our work is very much in line with \cite{he2016opponent},  where the authors build a general 
framework for modeling other agents in the reinforcement learning setting. 
\cite{he2016opponent} proposes a model that jointly learns a policy and the behavior of opponents 
by encoding observations of the opponent
into a DQN. Their Mixture of Experts architecture is able to discover different opponent strategy patterns in
two purely adversarial tasks. One difference between our work and \cite{he2016opponent}'s is that
we do not aim to infer other agents' strategies, but rather focus
on explicitly estimating their goals in the environment. 
Moreover, rather than using a hand designed featurization of the other agent's actions, in this work, the agent learns its model of the other end-to-end, based on its own model.  Another difference is that in this work, the agent runs an optimization to infer the other agent's hidden state, instead of inferring the other agent's hidden state via a feed-forward network.  In the experiments below, we show that SOM outperforms an adaptation of the method of \cite{he2016opponent} to our setting. 

\section{Experiments}
\label{experiments}

In this section, we evaluate our model SOM on three tasks:
\begin{itemize} 
\item The coin game, in Section \ref{coin}, which is a fully co-operative task where the agents' roles are symmetric.   
\item The recipe game, in Section \ref{crafting}, which is adversarial, but with symmetric roles.
\item The door game, in Section \ref{door}, which is fully cooperative but has asymmetric roles for the two players.
\end{itemize}
We compare SOM to three other baselines and to a model that has access to 
the ground truth of the other agent's goal. 
All the tasks considered are created in the Mazebase grid-world environment \cite{sukhbaatar2015mazebase}.
%We test our model on both cooperative (the agents have to maximize a shared reward) and competitive (the agents have to maximize an individual reward, and they cannot both accomplish their goals within the same episode).

% add sentence about the weight initialization

%%%%%%%%%%%%%%%%%%%%%%%%%%% BASELINES %%%%%%%%%%%%%%%%%%%%%%%%%%%%%%%%%%
\subsection{Baselines}
\label{baselines}

% TODO: explain baselines + explain why we did not directly compare to Hal's model or game

\textsc{True-Other-Goal (TOG)}: We provide an upper bound on the performance of our model given by 
a policy network which takes the other agent's \textbf{true} goal as input, $z_{other}$, as well as the state
features $s_{self}$ and its own goal $z_{self}$. Since this model has direct access to the true goal of the other agent,
it does not need a separate network to model the behavior of the other agent. The architecture of \textsc{TOG}
is the same as the one of \textsc{SOM}'s policy network, $f_{self}$.

% We provide an upper bound on the performance of our model given by 
% a model with the same architecture, which takes as input the ground truth of the other agent's hidden goal, 
% $z_{other}$, as well as the state features $s$ and its own goal $z_{self}$. 
% \rob{reword. need to make clear that we use $z_{other}$ rather than $\tilde{z}_{other}$} \rob{something about no need for opponent model in this case}.

% \rob{maybe use smallcaps for names of different methods. make latex macro. in particular, NoOtherModel seems a weird name. }. 
% \rob{change name, reword text} 

\textsc{No-Other-Model (NOM)}: The first baseline we use only takes as inputs the observation state $s_{self}$ and its own goal $z_{self}$. 
\textsc{NOM} has the same architecture as the one used for \textsc{SOM}'s policy network, $f_{self}$. This baseline has no
explicit model of the other agent or estimate of its goal.

% We also compare against a simple baseline, \textit{NoOtherModel}, which does not have any information about the other 
% agent's goal and makes no attempt to explicitly model the other agent or estimate its hidden state. 
% \textit{NoOtherModel} is an LSTM with the same architecture as the one used for SOM, taking as inputs $s$ and $z_{self}$.

\textsc{Integrated-Policy-Predictor (IPP)}: Starting with the architecture and inputs of \textsc{NOM}, we construct a stronger baseline, \textsc{IPP}, 
which has an additional final linear layer that outputs a probability distribution over the next action of the other agent.
Besides the A3C loss used to train the policy of this network, we also add a cross-entropy loss to train 
the prediction of the other agent's action, using observations of its behavior.

\textsc{Separate-Policy-Predictor (SPP)}: \citet{he2016opponent} propose an opponent modeling framework based on DQN. In their approach, a neural network (separate from the learned Q-network) is trained to predict the opponent's actions, given hand crafted state information specific to the opponent. An intermediate hidden representation from this network is given as input to the the Q-network. 

We adapt the model of \citet{he2016opponent} to our setting. In particular, we use A3C instead of DQN and we do not use the task-specific features used to represent the hidden state of the opponent.

The resulting model, \textsc{SPP},  consists of two {\em separate} networks, a policy network for deciding the agent's actions, 
and an opponent network for predicting the other's actions. The opponent network takes as input the state of the world $s$ and its own goal $z_{self}$, and outputs a 
probability distribution for the action taken by the other agent at the next step, as well as its hidden state
(given by the network's recurrence). As in IPP, we train the opponent policy predictor with a cross-entropy loss using the true actions of the other agent. At each step, the hidden state output by this network
is taken as input by the agent's policy network, along with the observation state and its own goal. Both the policy network and the opponent policy predictor are LSTMs with the same architecture as SOM.

In contrast to SOM, SPP does not explicitly infer the other agent's goal. Rather, it builds an implicit model of the opponent by predicting the agent's actions at each time step. In SOM, an inferred goal is given as additional input to the policy network. The analog of the inferred goal in SPP is the hidden representation obtained from the opponent policy predictor which is given as an additional input to the policy network.
% the two networks have different parameters and they do not take as input an estimate of the other agent's goal, so
% there is no explicit inference procedure.
% The opponent network takes as input the state of the world $s$ and its own goal $z_{self}$, and outputs a 
% probability distribution for the action taken by the other agent at the next step, as well as its hidden state
% (given by the network's recurrence). At each step, the hidden state outputted by this network
% is taken as input by the agent's policy network, along with the observation state and its own goal.
% \rob{possible figure to explain this?}
% Both networks are LSTMs with the same architecture as the one used for SOM. 

%As we have mentioned, this is a similar version to the method designed by \cite{he2016opponent}, 
%with the caveat that it is adapted to work with A3C and the opponent representation used to decide the agent's actions
%are not hand-crafted, but rather learned using supervision of the opponent's actions. 

\textbf{Training Details.} In all our experiments, we train the agents' 
policies using A3C \cite{mnih2016asynchronous}
with an entropy coefficient of 0.01, a value loss coefficient of 0.5,
and a discount factor of 0.99. The parameters of the agents' policies are optimized
using Adam \cite{kingma2014adam} with $\beta_1 = 0.9$, $\beta_2 = 0.999$, $\epsilon = 1 \times 10^{-8}$, 
and weight decay 0. 
SGD with a learning rate of 0.1 was used for inferring the other agent's goal, $\tilde{z}_{other}$.

%  \rob{of policy network?}
The hidden layer dimension of the policy network was 64 for the Coin and Recipe Games and 128 for the Door Game.
We use a learning rate of $1 \times 10^{-4}$ for all the games and models. 

% The observation state $s$ of each agent is represented by few-hot vectors, 
% indicating the relative locations of all other objects in the environment (including the other player).

The observation state $s$ is represented by few-hot vectors indicating 
the locations of all the objects in the environment, as well as the locations of the self and the other.
The dimension of this input state is $1\times nfeatures$, 
where the number of features is 384, 192, and 900
for the Coin, Recipe, and Door games, respectively.

% \rob{we also need to discuss input representation of mazebase}
% \rob{need to say something about 5 different random seeds for each experiment}
% XXX: add the different things we are trying to probe in our different environments

For each experiment, we trained the models using 5 different random seeds.
All the results shown are for 10 optimization updates of $\tilde{z}_{other}$
at each step in the game, unless mentioned otherwise. 
% ,
% and we report the corresponding mean and standard deviation of the reward or
% success ratio. 

% For all the experiments, we report the mean and standard deviation of the reward
% across 5 runs with different random seeds.

% %%%%%%%%%%%%%%%%%%%%%%%%%%% COOPERATIVE %%%%%%%%%%%%%%%%%%%%%%%%%%%%%%%%%%
% \subsection{Cooperative Domains}
% \label{collaborative}

% TODO: intro to the different types of things we are testing with the different coop games

%%%%%%%%%%%%%%%%%%%%%%%%%%% COIN GAME %%%%%%%%%%%%%%%%%%%%%%%%%%%%%%%%%%

% \begin{figure}[htbp]
% 	    \centerline{%
% 		    \includegraphics[width=0.5\columnwidth]{diagrams/coin_diagram.pdf}%
% 		}%
% 		\caption{Illustration of the coin game. \rob{try and get this to appear near text explaining game}}
% 		\label{model}
% \end{figure}

\begin{figure*}[ht]
    \begin{center}

	    \centerline{%
		    \includegraphics[width=0.2\linewidth]{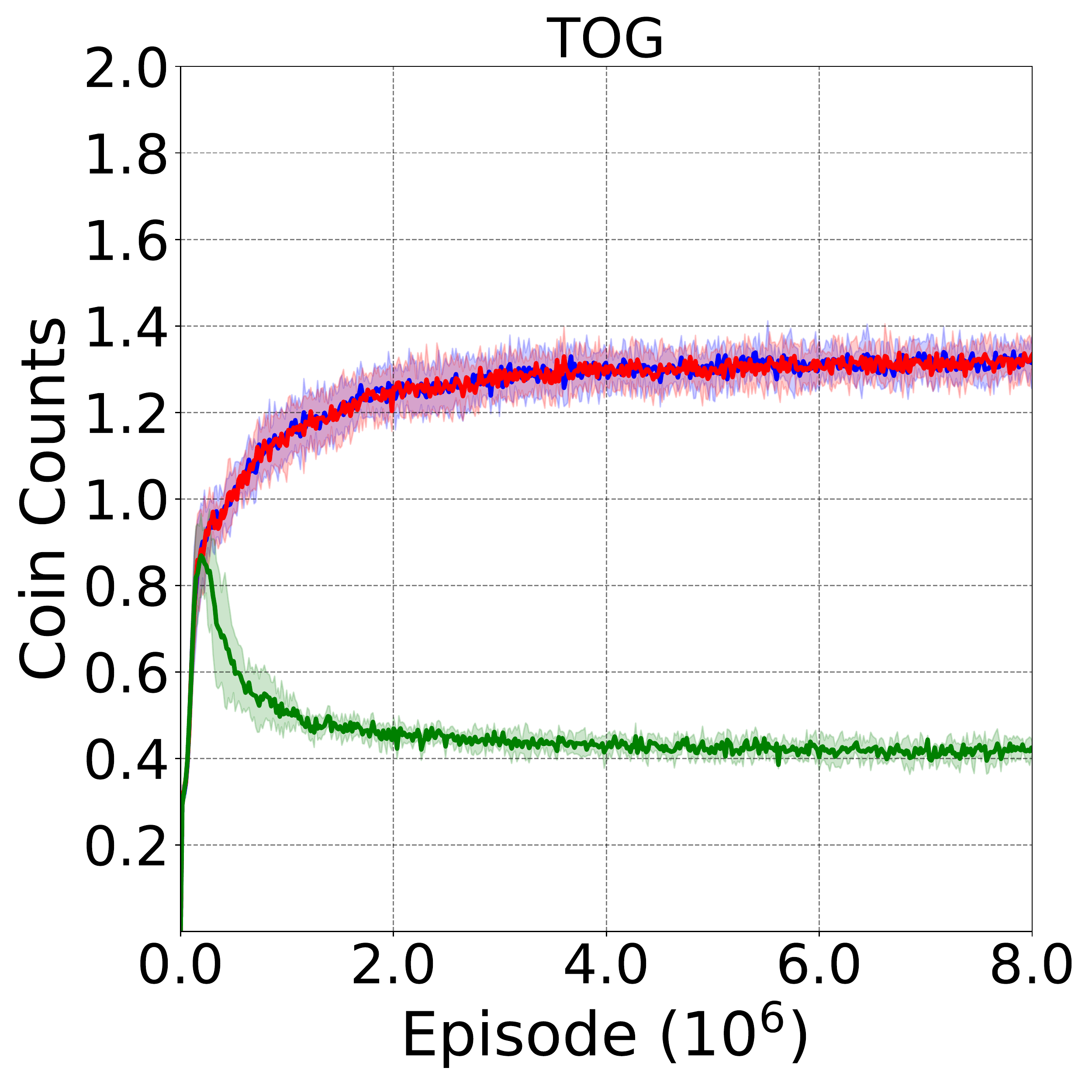}%
		    \includegraphics[width=0.2\linewidth]{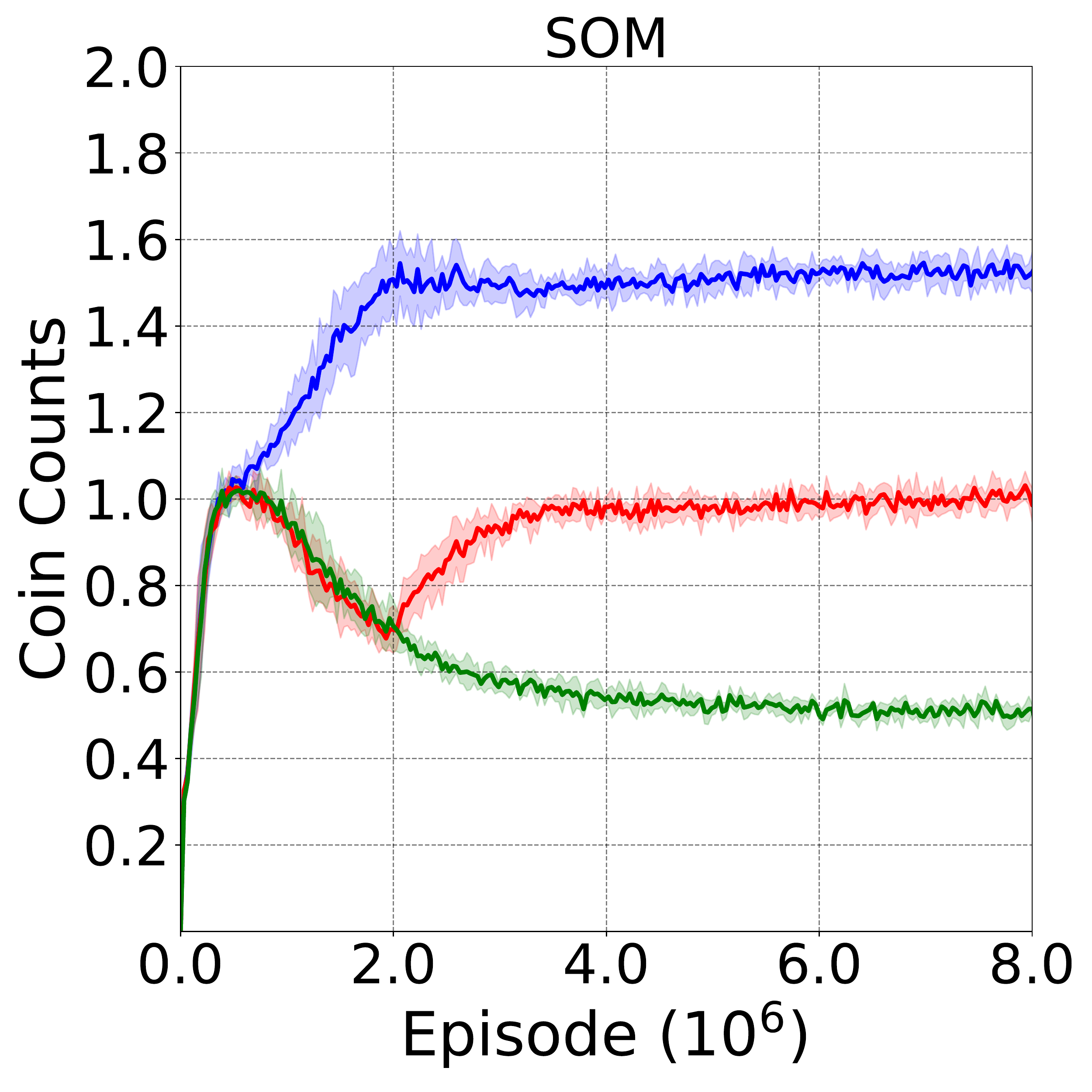}%
		    \includegraphics[width=0.2\linewidth]{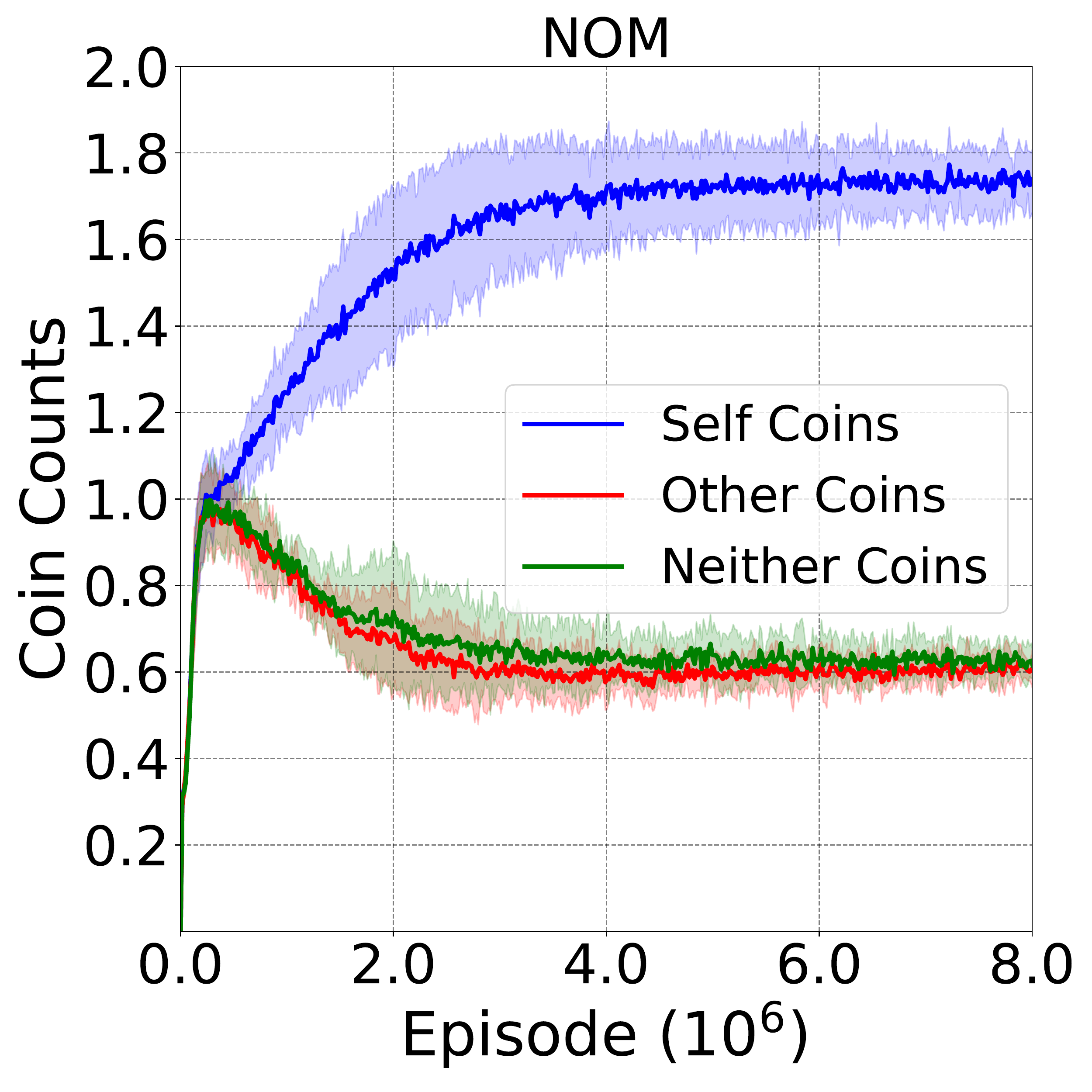}%
		    \includegraphics[width=0.2\linewidth]{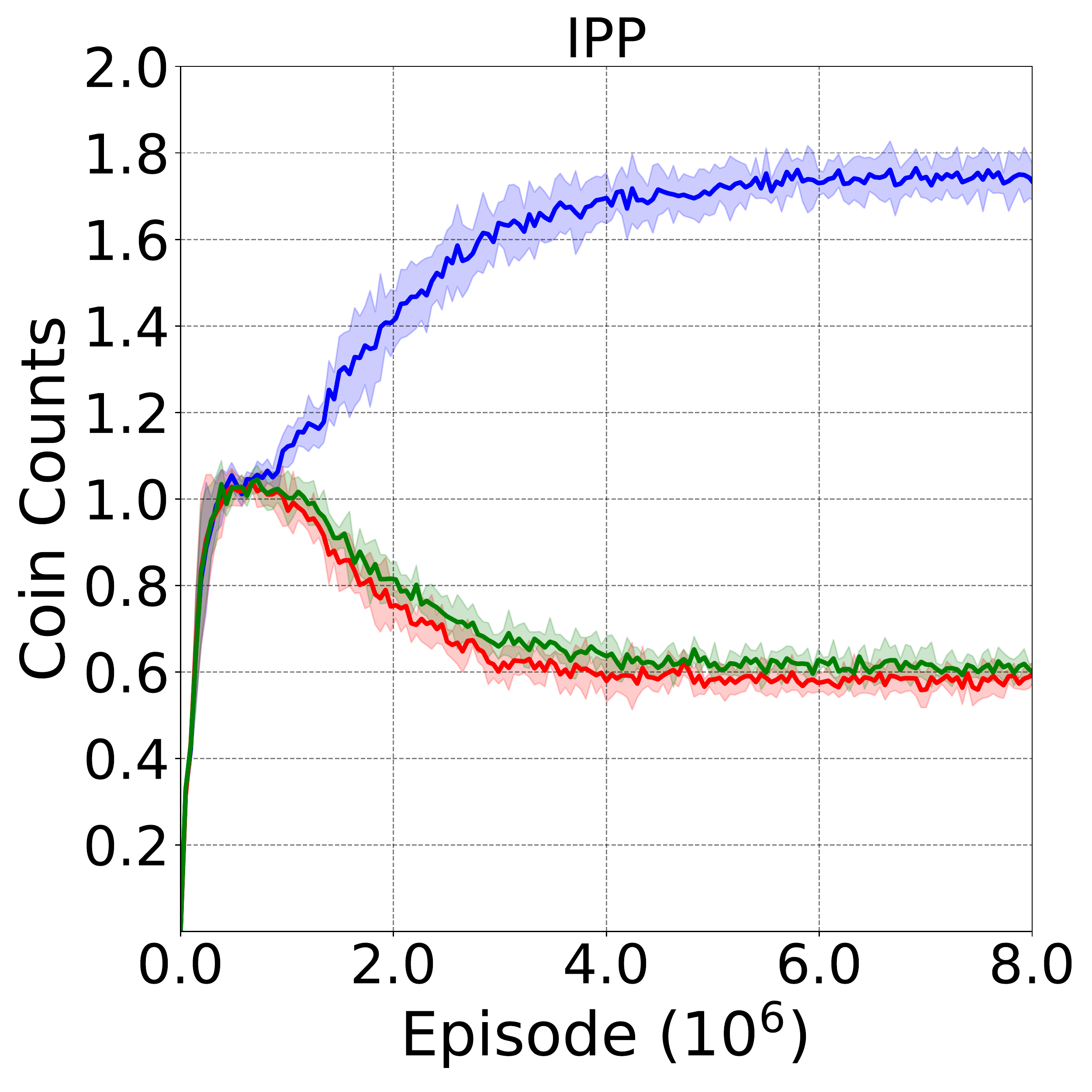}%		   
		    \includegraphics[width=0.2\linewidth]{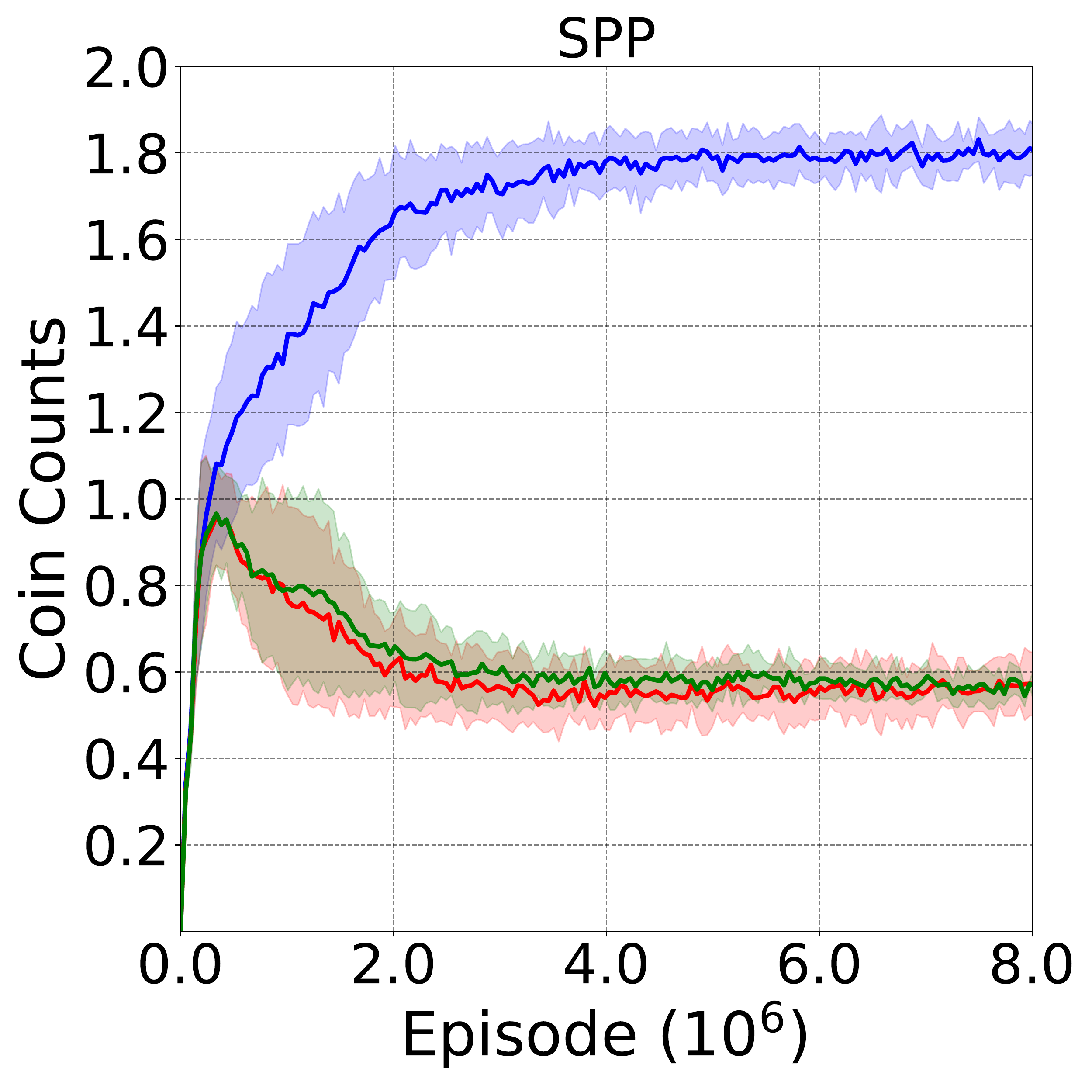}%		   
		}%
        \caption{\textbf{Coin Strategy:}  Average number of collected coins per episode corresponding to the color of the Self (blue), Other (red), or Neither (green) by the agents using TOG (left), SOM (center-left), NOM (center), IPP (center-right), and SPP (right). 
        The optimal strategy is to pick up as many Self as Other coins on average, across a number of episodes, and no Neither coins. Being able to collect more Other than Neither coins indicates
        that the agent is able to accurately infer the other agent's color early enough during some of the episodes and uses this information to collect
        more Other, instead of Neither coins, which increases its reward.
        The TOG model learns to collect just as many Self as Other coins, while all the baselines only learn to collect more Self coins, but cannot distinguish between the Other and Neither coins. SOM learns to collect significantly more Other coins than Neither. This shows that SOM converges to a closer-to-optimal strategy using its guess of the other's goal.}  
        % \vspace{-2em}
		\label{fig:coin_strategy}
	\end{center}
\end{figure*}

% \rob{can we put labels above each subplot to indicate the method. much easier than reading text}.
%\rob{need to be clear what should happen -- put 1 setnence in here explanining that before going into details of individual methods}. 

% \roberta{added text below explaining results and refs to figs. comments/edits?}

\subsection{Coin Game.}
\label{coin}
\begin{figure}[htbp]
	    \centerline{%
		    \includegraphics[width=0.3\columnwidth]{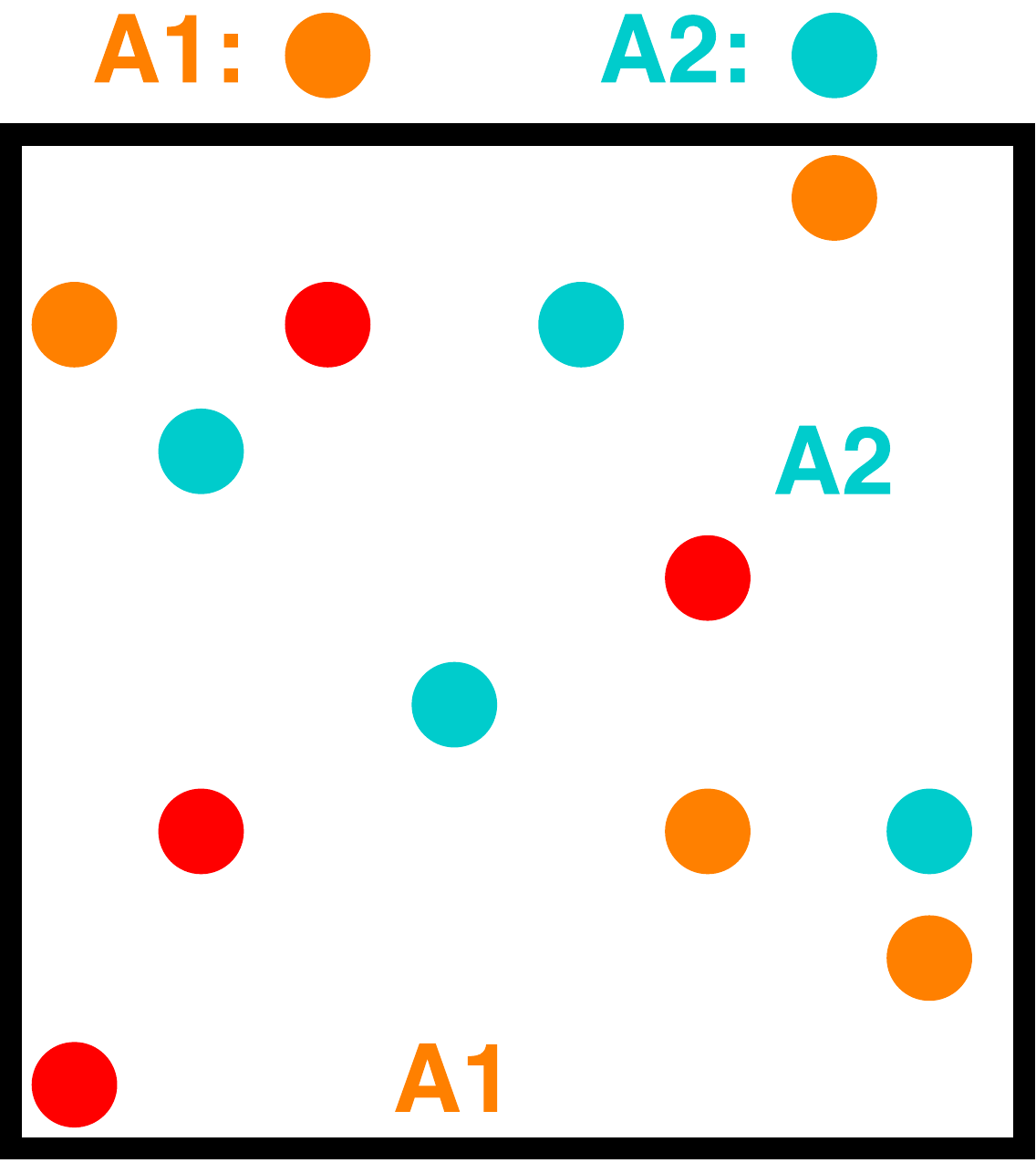}
		    \vrule
		    \includegraphics[width=0.3\columnwidth]{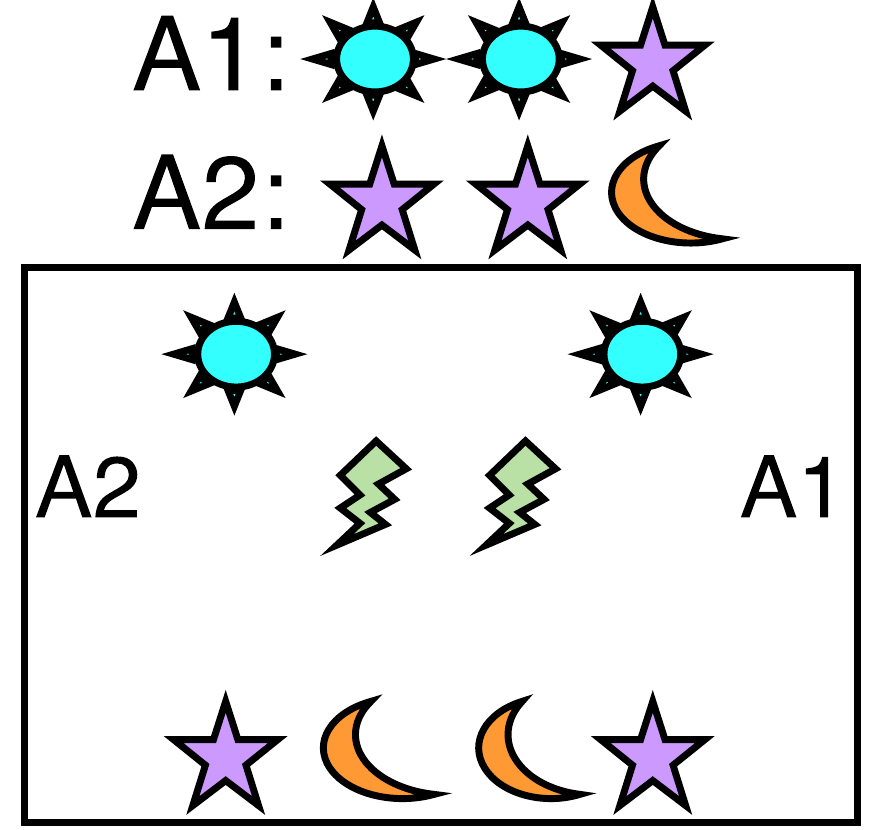}%
		    \vrule
		    \includegraphics[width=0.3\columnwidth]{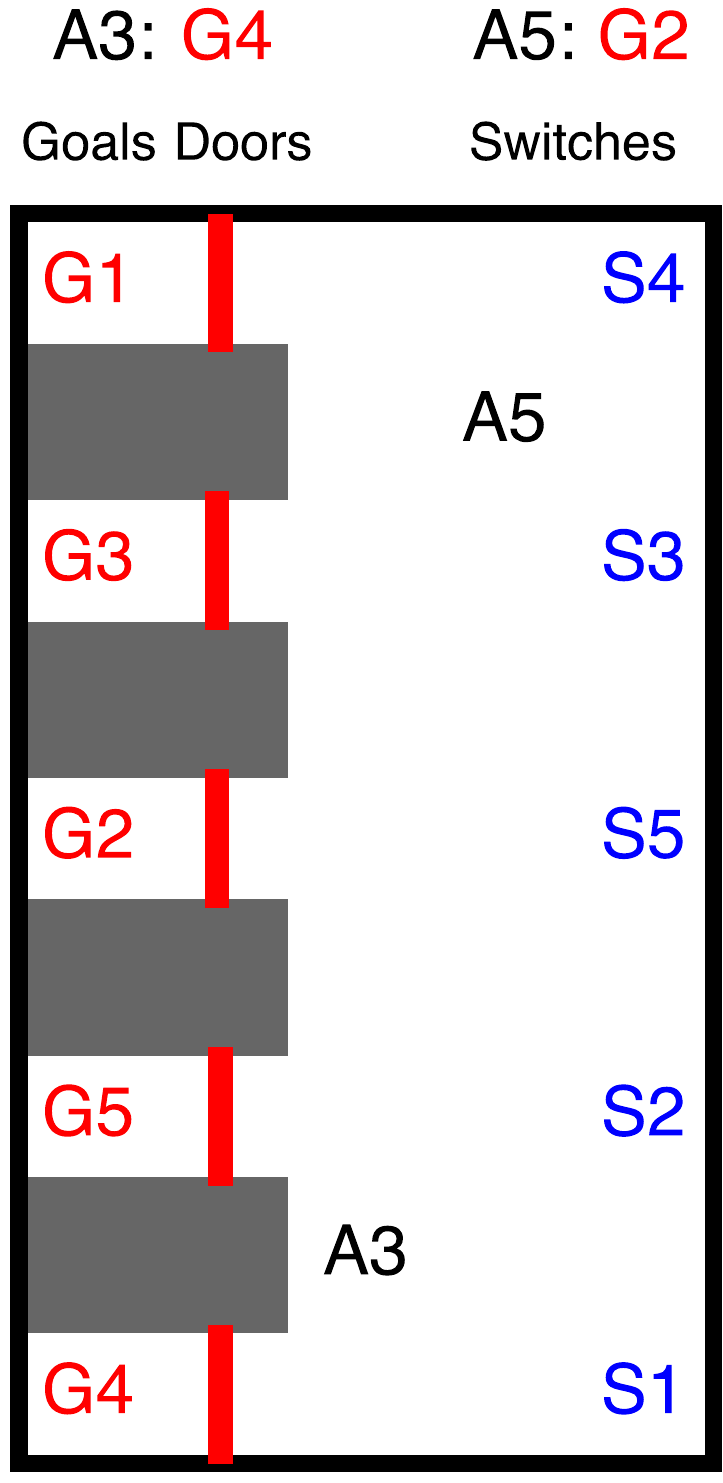}%
		}%
		\caption{Illustration of the Coin (left), Recipe (center), and Door (right) games
		Above each ones we show the agents' goals (not visible to one another).}
        % \vspace{-1.5em}
		\label{fig:game_diagrams}
\end{figure}

First, we evaluate the model on a fully cooperative task, in which the agents 
can gain more reward when using both of their goals rather than only their own goal.
So it is in the best interest of each agent to estimate the other player's goal and 
use that information when taking actions.
The game, shown in the left diagram of Figure~\ref{fig:game_diagrams}, takes place on a $8\times8$ grid containing 12 coins of 3 different colors (4 coins of each color). 
At the beginning of each episode, the agents are randomly assigned one of the three colors. 
The action space consists of: go up, down, left, right, or pass.
% \rob{something about action space: l,r,u,d, i.e. no pickup} 
Once an agent steps on a coin, that coin disappears from the grid. The game ends after 20 steps (i.e. each agent takes 10 steps).
The reward received by both agents at the end of the game %is the sum of the squared number of coins of the same color, for the colors matching the two players' goals, minus the squared number of coins of the third color (the one which is not a goal for any of the two agents):
is given by the formula below:
% \vspace{-1.5em}
\begin{align*}\label{eq:reward_coin}
R &= (n_{C_{self}}^{self} + n_{C_{self}}^{other})^2 + (n_{C_{other}}^{self} + n_{C_{other}}^{other})^2  \\
  &- (n_{C_{neither}}^{self} + n_{C_{neither}}^{other})^2,
\end{align*}
% \vspace{-1.5em}
where $n_{C_{self}}^{other}$ is the number of coins of the self's goal-color, which 
were collected by the other agents, 
and $n_{C_{neither}}^{self}$ is the number of coins corresponding to neither
of the agents' goals, collected by the self. 
For the example in Figure~\ref{fig:game_diagrams}, 
agent 1 has $C_{self} = $ orange and $C_{other} = $ cyan,
while agent 2's $C_{self}$ is cyan and $C_{other}$ is orange. $C_{neither}$ is red for both agents.
% \rob{Maybe use example from Fig.4 to make clearer}

The role of the penalty for collecting
coins that do not correspond to any of the agents' goals is to avoid convergence to
a brute force policy in which the agents can gain a non-negligible amount of reward
by collecting all the coins in their proximity, without any regard to their color.

\begin{figure}[ht]
    \begin{center}
	    \centerline{%
		    \includegraphics[width=0.75\columnwidth]{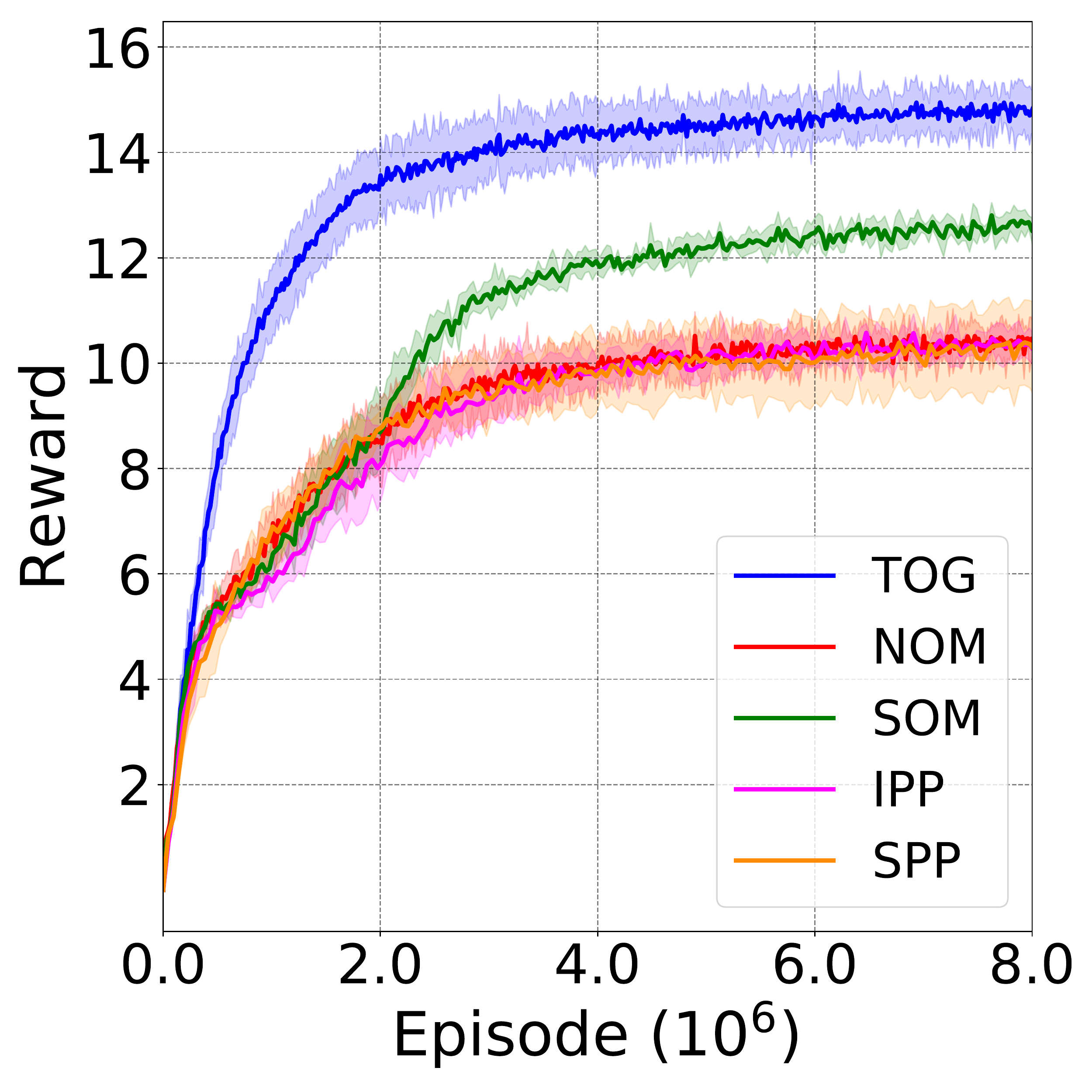}%
		}%
        \caption{\textbf{Coin Performance:} Average reward obtained on the Coin game by SOM (green), TOG (blue), NOM (red), 
        IPP (magenta), and SPP (orange). SOM performs better than all the baselines.} 
        % \vspace{-3.0em}
		\label{fig:coin_reward}
	\end{center}
\end{figure}

To maximize its return, each agent needs to collect coins of its own or its collaborator's color, but not those of the remaining color.
%In particular, this reward structure incentivizes the agents to collectively pick more coins matching only one of their two goal-colors instead of each of them picking more coins matching its own goal-color. 
Thus, when both agents are able to infer their collaborators' goals with high accuracy
and as early as possible in the game, they can use that information to maximize their shared utility.
% Thus, being able to accurately and rapidly infer the other agent's hidden goal-color, 
% would help the agents gain more reward. 

Figure~\ref{fig:coin_reward} shows the mean and standard deviation of the reward
across 5 runs with different random seeds obtained by SOM. Our model 
clearly outperforms all the baselines on this task.
%, including SPP (which is a close approximation of the method from \cite{he2016opponent}).
We also show the empirical upper bound on the reward using the model which takes as input
the true color assigned to the other agent.

Figure~\ref{fig:coin_strategy} analyzes the strategies of the different models by looking at the proportion of coins of each type collected by the agents. The optimal strategy is for each agent to maximize $n_{C_{self}}^{self} + n_{C_{other}}^{self}$ and $n_{C_{neither}}^{self} = 0$. Due to the randomization of objects in the environment, this amounts to each agent collecting an equal number of coins of its own color and coins of the other's color on average, across a large number of episodes (i.e. $\bar{n}_{C_{self}}^{self} = \bar{n}_{C_{other}}^{self}$). 

%The optimal strategy for this game is one in which each agent collects on average an equal number of Other Coins (i.e. coins corresponding to the other agent's assigned goal-color) and Self Coins (i.e. coins corresponding to its own goal-color). To maximize return,  the agent should avoid collecting coins of the remaining color which was not assigned as a goal to any of the two players (referred to as Neither Coins).
Indeed, this is the strategy learned by the model with perfect information of the other agent's goal (TOG).
SOM also learns to collect significantly more Other than Neither coins (although not as many as Self coins), indicating its ability to distinguish between the two types, at least during some of the episodes. This means that SOM can accurately infer the other agent's goal early enough during the episode
and use that information to collect more Other Coins, thus gaining more reward than if it were only using
its own goal to direct its actions. 

In contrast, the agents trained with the three baseline models collect significantly more Self coins, and
as many Other as Neither coins on average. This shows that they learn to use their own goal for 
gaining reward, but they are unable to use the hidden goal of the other agent for further increasing their reward. 
Even if IPP and SPP are able to predict the actions of the other player with an accuracy of about $50\%$, they 
do not learn to distinguish between the coins that would increase (Other) and those
that would decrease (Neither) their reward. 
This shows the weaknesses 
of using an implicit model of the other agent to maximize reward on certain tasks.

%\roberta{titles for the games? or small captions below them?}

% TODO: explain crafting game
% TODO: explain pretraining stage
\subsection{Recipe Game.}
\label{crafting}
Agents in adversarial scenarios can also benefit from having a model of their
opponents, which would enable them to exploit the weaknesses of certain players. 
With this motivation in mind, we evaluate our model on a game in which the agents have to craft certain 
compositional recipes, each containing multiple items found in the environment. 
The agents are given as input the names of their goal-recipes, 
without the corresponding components needed to make it. 
The resources in the environment are scarce, so only one of the agents can 
craft its recipe within one episode.

\begin{figure*}[ht]

	    \centerline{%
		    \includegraphics[width=\linewidth]{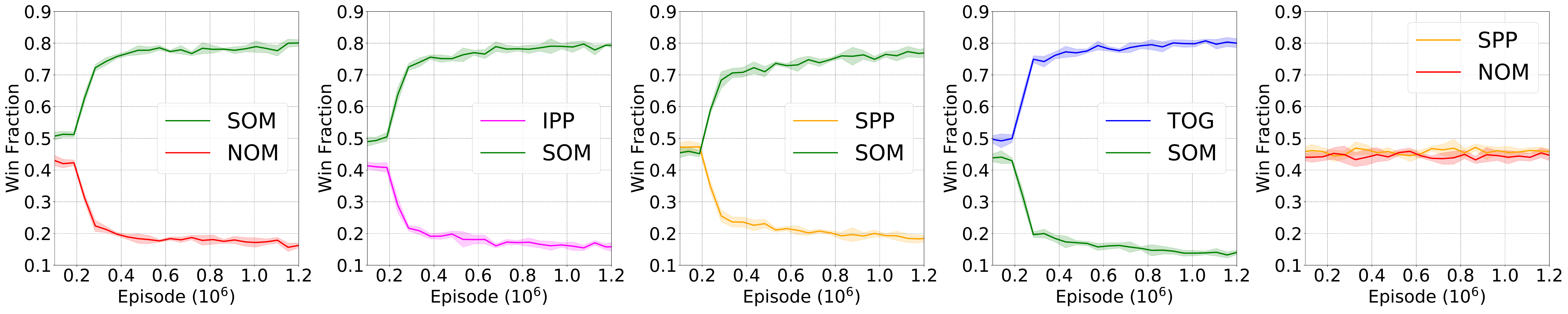}%
		}%
        \caption{\textbf{Recipe Performance:} Average fraction of success in the Recipe game by SOM-NOM (left), SOM-IPP (center-left), SOM-SPP (center-center), SOM-TOG (center-right), 
        NOM-AcrPredSep (right). The plots show the performance of SOM with 5 optimization updates of $\tilde{z}_{other}$
        at each step in the game.}  
        % \vspace{-1em}
        \label{fig:crafting_win}
\end{figure*}

%\rob{this is confusing. maybe just give all 4 receipes}. 
As illustrated in Figure~\ref{fig:game_diagrams} (center),
there are 4 types of items: \{sun, star, moon, lightning\} 
and 4 recipes: \{sun, sun, star\}; \{star, star, moon\}; 
\{moon, moon, lightning\}; \{lightning, lightning, sun\}. 
The game is played in a $4\times6$ grid, which contains 8 items in total, 2 of each type.

% as well as 4 possible recipes, each requiring 3 different items. 
% Each recipe requires two items of the same type and a third item of a different type 
% (e.g., recipe 1: \{sun, sun, star\}; recipe 2: \{star, star, moon\}; 
% recipe 3: \{moon, moon, lightning\}; recipe 4: \{lightning, lightning, sun\} etc.).
% as illustrated in Figure~\ref{fig:game_diagrams}.
% \rob{refer to fig}

At the beginning of each episode, we randomly assign a recipe to one of the agents,
and then we randomly pick a recipe for the other agent so that it has overlapping items 
with the recipe of the first agent. 
%Hence, in each game, the agents need a total of 3 items of a certain  type (the type needed for both of their recipes) to make their recipes, but crucially there are only 2 items of that type in the environment. 
This ensures that the agents are competing
for resources within each episode. At the end of the episode, each agent receives a reward of +1
for crafting its own recipe and a penalty of -0.1 for each item it picked up not needed for making its recipe.

We designed the layout of the grid so that neither agent has an initial advantage by being closer to the
scarce resource. At the beginning of each episode,
%the agents and objects are symmetrically placed with respect to the vertical axis going through the middle of the grid. 
%More explicitly, 
one of the agents starts on the left-most column of the grid, 
while the other one starts on the right-most column, at the same y-coordinate. 
Their initial y-coordinate as well as which agent starts on the left/right is randomized. 
Similarly, one item of each of the 4 different types is placed at random in the grid formed by
the second and third columns of the maze, from left to right. The rest of the items
are placed in the forth and fifth columns, so that the symmetry with respect to the vertical axis
is preserved (i.e. items of the same type are placed at the same y-coordinate, and symmetric x-coordinates).

Agents have six actions to choose from: pass, go up, down, left, right, or pick 
(for picking an item, which then disappears from the grid). 
The first agent to take an action is randomized. The game ends after 50 steps.
%\arthur{I thought the actions were simultaneous?}

%Besides learning the decomposition of each recipe into its items, the agent must also infer its opponent's recipe so that it prioritizes collecting the scarce resource first (i.e. the item which is part of both of their recipes).

We pretrain all the baselines on a version of the game which does not have overlapping recipes, in order
to ensure that all the models learn to pick up the corresponding items, given a recipe as goal.
All of the models learn to craft their assigned recipes $\sim90\%$ of the time on this simpler task. 
Then, we continue training the models on the adversarial task in which their recipes
overlap in each episode. SOM is initialized with a pretrained NOM network.  

Figure~\ref{fig:crafting_win} shows the winning fraction for different pairs played against each other in the Recipe game. 
For the first 100k episodes, the models are not being trained. We can see that SOM significantly outperfroms NOM, IPP, and SPP,
winning $\sim75-80\%$ of the time, while the baselines can only win $\sim15-20\%$ of the games.
SPP ties against NOM, and TOG outperforms SOM by a large margin. We also played the same types of agents against each other and
they all win $\sim40-50\%$ of the games.

% \rob{Need to refer to figs and discuss results}

%%%%%%%%%%%%%%%%%%%%%%%%%%% DOOR GAME %%%%%%%%%%%%%%%%%%%%%%%%%%%%%%%%%%
% \begin{figure}[htbp]
% 	    \centerline{%
% 		    \includegraphics[width=0.5\columnwidth]{diagrams/door_diagram_v3.pdf}%
% 		}%
% 		\caption{Illustration of the door game.}
% 		\label{model}
% \end{figure}

% TODO: explain why this is a more challenging task of our model...
% TODO: ...by introducing a potential limitation of our model and combat this with the Results on the Door Game
% TODO: explain the idea of the pool of agents
\subsection{Door Game.}
\label{door}

%One of the main assumptions made by our approach is that the other player's policy is similar to your own, when conditioned on the goals of the two.   
In this section, we show that on a collaborative task with asymmetric roles and multiple possible partners, the agents can learn to figure out what role they should be playing in each game based on their partners actions.  

%However, this may seem too restrictive since there are many scenarios one can imagine in which the two agents must play different roles for solving some task, which translates in different policies for the same goal.
%In particular, many collaborative tasks require the agents to solve different pieces of the puzzle for the two to get reward. 
%One can suspect that such asymmetries in the task would lead SOM to be ineffective at predicting other players' behavior and inferring their intentions using its own policy.  

\begin{figure}[ht]
    \begin{center}
	    \centerline{%
		    \includegraphics[width=0.75\columnwidth]{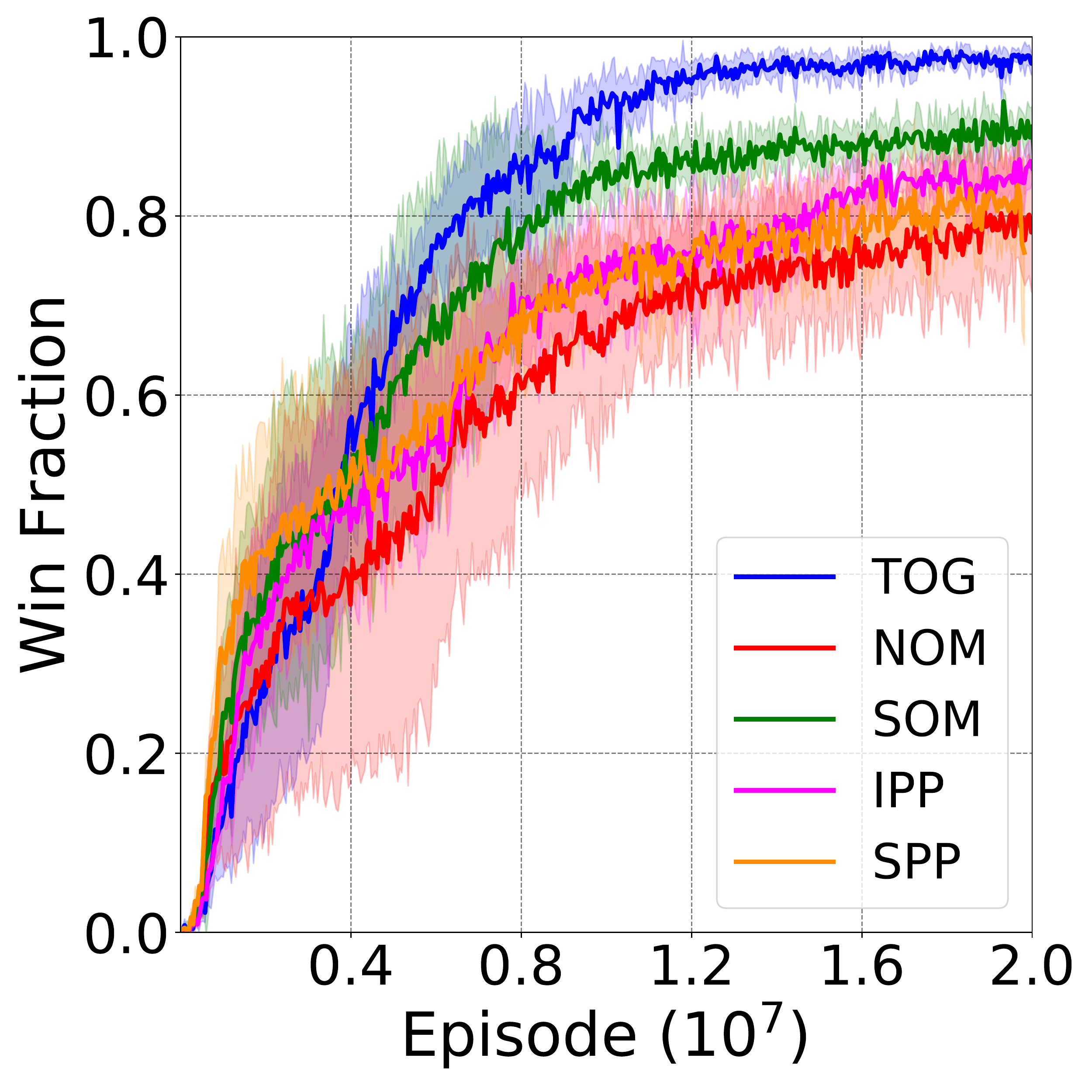}%
		}%
        \caption{\textbf{Door Performance:} Average fraction of success on the Door game by SOM 
        (green), TOG (blue), NOM (red), 
        IPP (magenta), and SPP (orange). On average, SOM performs better than all the baselines.} % \arthur{I wish the plot  went farther to the right, maybe to $2\times10^7$ episodes.}} 
        % \vspace{-3.5em}
		\label{fig:door_reward}
	\end{center}
\end{figure}

%For the purpose of probing our method on a task with asymmetric optimal policies for the same goal, we constructed the following game, in which
In the Door game, two agents are located in a $5\times9$ grid, 
with 5 goals behind 5 doors on the left wall, and 5 switches on the right wall of the grid. 
The game starts with the two players in  random squares on the grid, 
except for the ones occupied by the goals, doors, or switches, we illustrated in Figure~\ref{fig:game_diagrams}.
% \rob{refer to fig}
Agents can take any of the five actions: go up, down, left, right or pass. 
An action is invalid if it moves the player outside of the border or to
a square occupied by a block or closed door. 
Both agents receive +3 reward when either one of them steps on its goal and they are penalized -0.1 for each step they take. 
The game ends when one of them gets to its goal or after 22 steps. 
All the goals are behind doors which are open only as 
long as one of the agents sits on the corresponding switch for that door. 

At the beginning of an episode, each of the two players is randomly selected from a pool of 5 agents 
% \rob{we are going with a fixed number? need to make clear agents don't know who others are and they have different policies.}
and receives as input a random number from 1 to 5 corresponding to its goal. 
Each of the 5 agents has its own policy which gets updated at the end of each episode they play. 
%whenever they play an episode. 
Note that the agents' identities are not visible (i.e. there is no indication in the state features that
specifies the id's of the agents playing during a given episode). 
This restriction is important in order to ensure that the agents cannot gain advantage 
by specializing into the two roles needed to win (i.e. goal-goer and switch-puller) 
and identifying the specialization of the other player by simply observing its unique id.

The agents need to cooperate in order to receive reward.
In contrast to our previous tasks, the two players must take different roles. 
In fact, the player who sits on the switch should ignore its own goal and instead infer the other's goal,
while the player who goes to its goal does not need to infer the other's goal, but only use its own. 
In order to 
sit on the correct switch, an agent has to infer the other player's goal from their observed 
actions. 
%The two players need to solve different subtasks in order to win the game: one of them has to go to its goal, while the other one has to sit on the switch corresponding to its collaborator's goal.
The only way in which an agent can use its own policy to model the other player is if each agent
learns to play both roles of the game, i.e. go to its own goal and also open its collaborator's door 
by sitting on the corresponding switch. Indeed, we see that the agents learn to play both roles
and they are able to use their own policies to infer the other player's goals when needed.

Fig~\ref{fig:door_reward} shows the mean and standard deviation of the winning fraction obtained
by one of the agents on the Door game. While our model is still able to outperform the three baselines, 
the gap between the performance of our model and that of IPP or SPP (an approximate version of \cite{he2016opponent}) 
is smaller than in the previous tasks. 
However, this is a more difficult task for our model since it needs the agents to learn performing
both roles before effectively use its own policy to infer the other agent's goal. 
Nevertheless, we see that SOM training allows the agents to play both roles in an asymmetric cooperative game, and to infer the goal and role of the other player.

% \roberta{not sure if the below para is clear}
%The fact that SOM is able to do slightly better than these other methods on this particular taskindicates that its use is not limited to symmetrically-structured scenarios in which all the agents have the same policy, given a goal. 

% TODO: add sentence about how the performance compares to the baselines and why etc.

% %%%%%%%%%%%%%%%%%%%%%%%%%%% ADVERSARIAL %%%%%%%%%%%%%%%%%%%%%%%%%%%%%%%%%%
% \subsection{Adversarial Domains}
% \label{adversarial}

% %%%%%%%%%%%%%%%%%%%%%%%%%%% RESULTS %%%%%%%%%%%%%%%%%%%%%%%%%%%%%%%%%%
% \section{Results}
% \label{results}
\subsection{Analyzing the goal inference}

In this section we further analyze the ability of the SOM models to infer other's intended goals.

% \begin{figure*}[ht]
%     \begin{center}

% 	    \centerline{%
% 		    \includegraphics[width=0.3\linewidth]{figures/coins_percent_inf_step.pdf}%
% 		    \includegraphics[width=0.3\linewidth]{figures/recipes_percent_inf_step.pdf}%
% 		    \includegraphics[width=0.3\linewidth]{figures/doors_percent_inf_step.pdf}%
% 		}%
%         \caption{\textbf{Inference Accuracy during an Episode:} Percentage of episodes in which the agent's estimate of the other's goal at that
%         time step is correct for the Coin (left), Crafting (center), and Door (right) games, as a function of step in the game. }  
% 		\label{fig:coin_strategy}
% 	\end{center}
% \end{figure*}

\begin{figure}[ht]
    \begin{center}

	    \centerline{%
		    \includegraphics[width=0.33\linewidth]{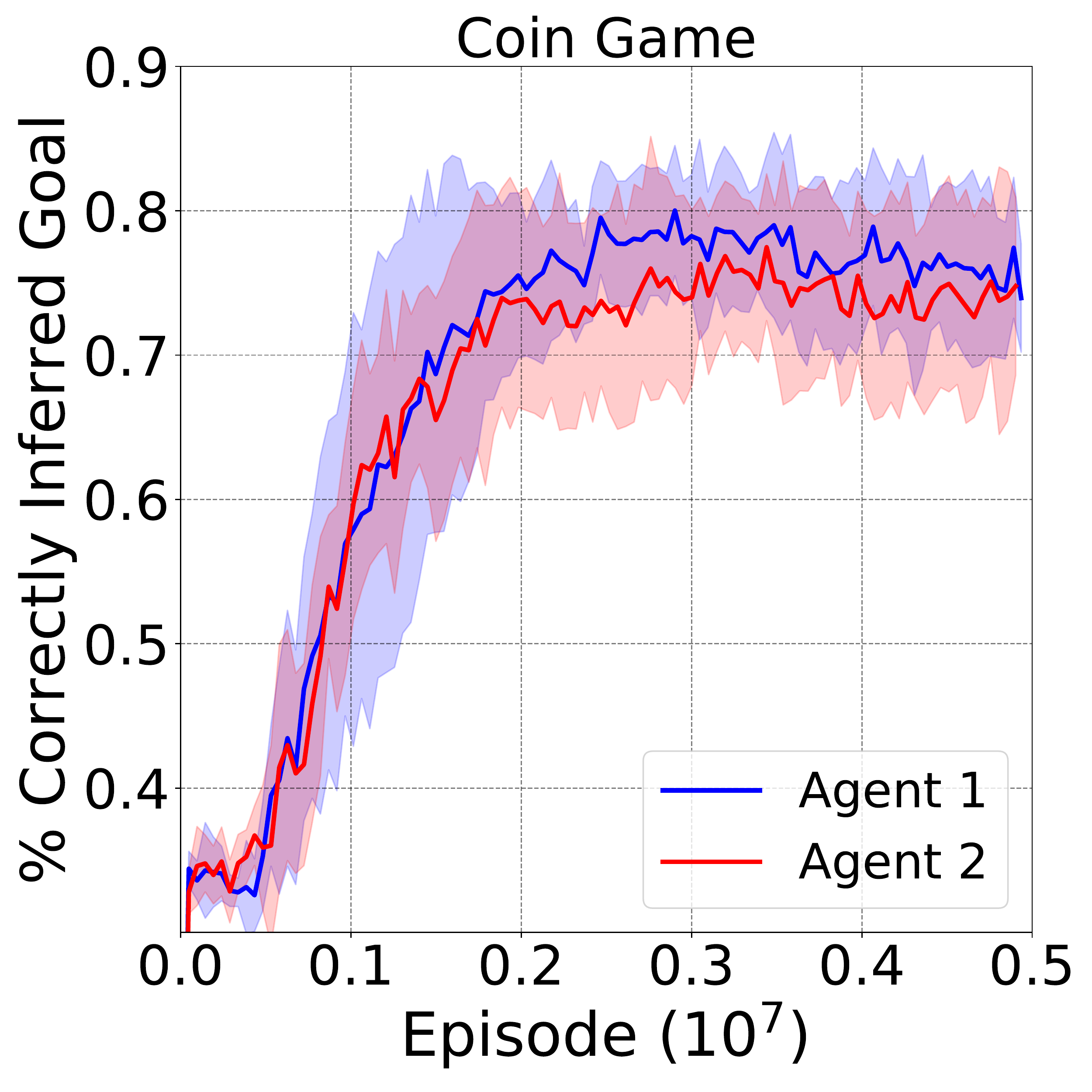}%
		    \includegraphics[width=0.33\linewidth]{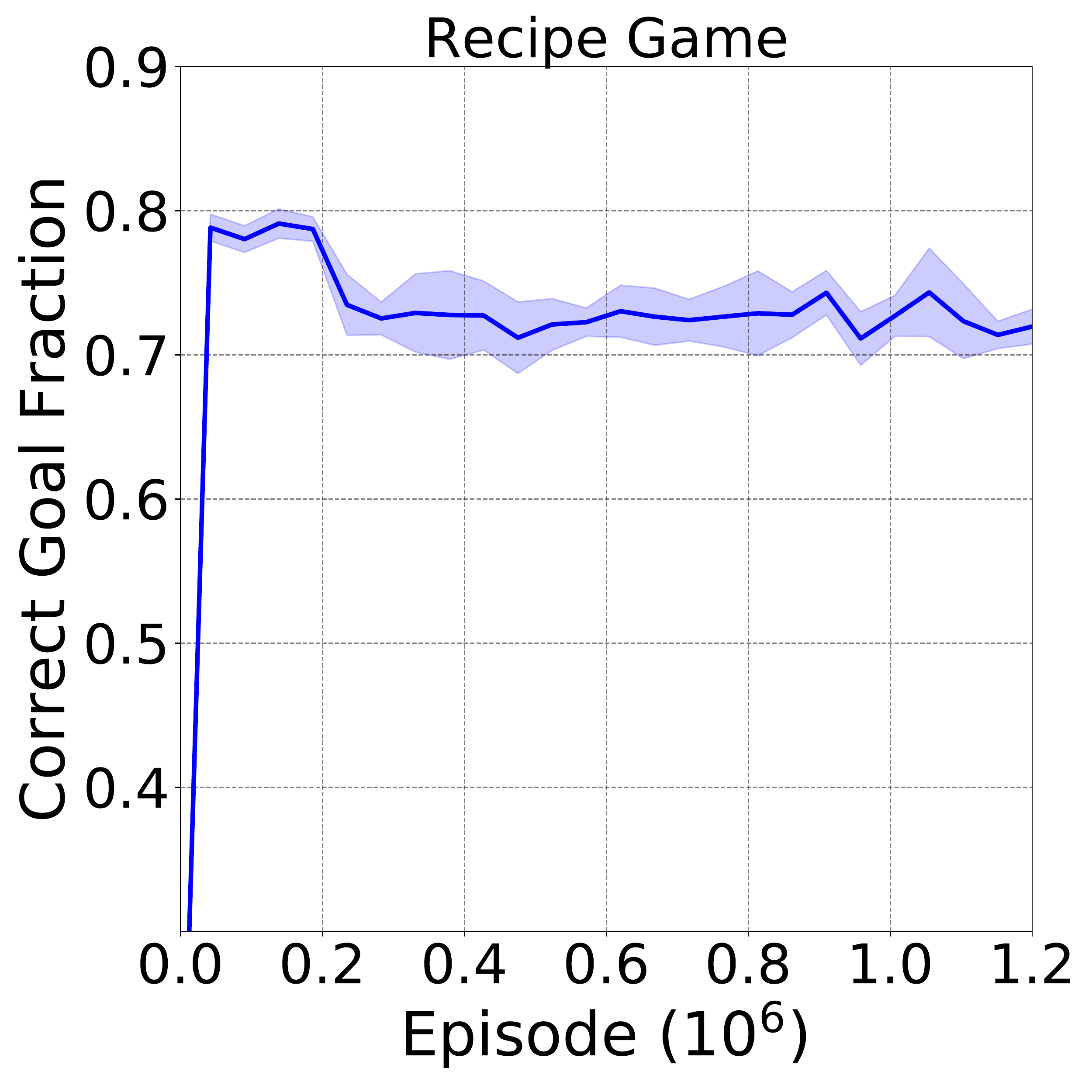}%
		    \includegraphics[width=0.33\linewidth]{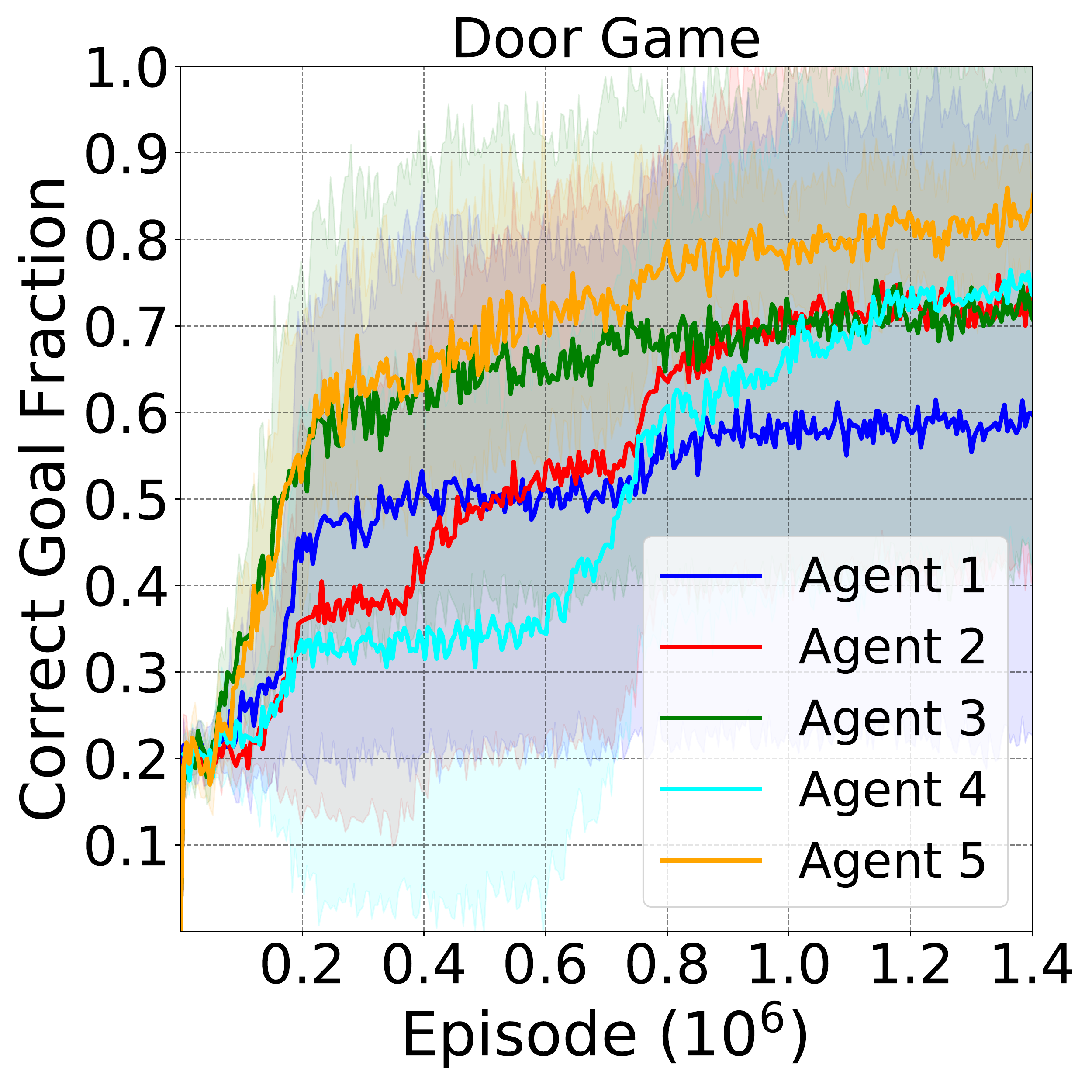}%
		}%
        \caption{\textbf{Inference Accuracy during Training:} The mean fraction of episodes in which the agent correctly 
        infers the other's goal for the Coin (left), Recipe (center), and Door (right) games, as a function of training epoch. The estimate of the other's goal is considered correct if it remains accurate during all the following steps in the game.}
% 		\vspace{-2em}
		\label{fig:inf_acc_train}
	\end{center}
\end{figure}

Figure~\ref{fig:inf_acc_train} shows the fraction of episodes in which the goal of the other agent is correctly inferred. 
We consider that the goal is correctly inferred only when the estimate of the other's goal remains accurate until the end of the game,
so that we avoid counting the episodes in which the agent might infer the correct goal by chance at some intermediate step in the game. 
In all the games, the SOM agent learns to infer the other player's goal with a mean accuracy ranging from 
$\sim 60 - 80\%$. Comparing the second plot in Figure~\ref{fig:coin_strategy} with the left plot in Figure~\ref{fig:inf_acc_train}, one can observe
that the SOM agent starts distinguishing Other from Neither coins after approximately 2M training epochs, which coincides with the time 
when the mean accuracy of the inferred goal converges to $\sim 75\%$. The Door Game (right) presents higher variance since the agents learn to 
use and infer the other's goal at different stages during training.

\begin{figure}[ht]
    \begin{center}

	    \centerline{%
		    \includegraphics[width=0.33\linewidth]{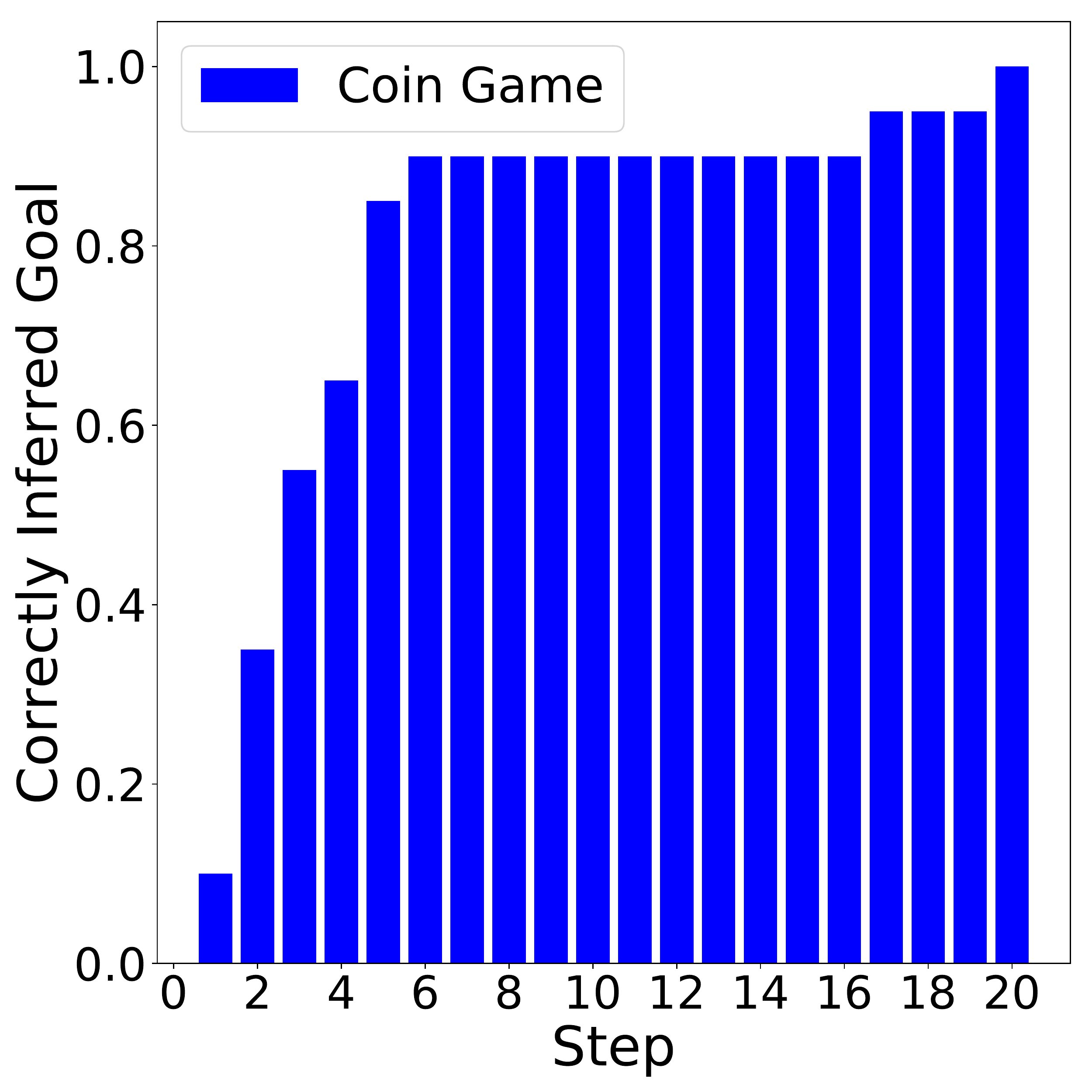}%
		    \includegraphics[width=0.33\linewidth]{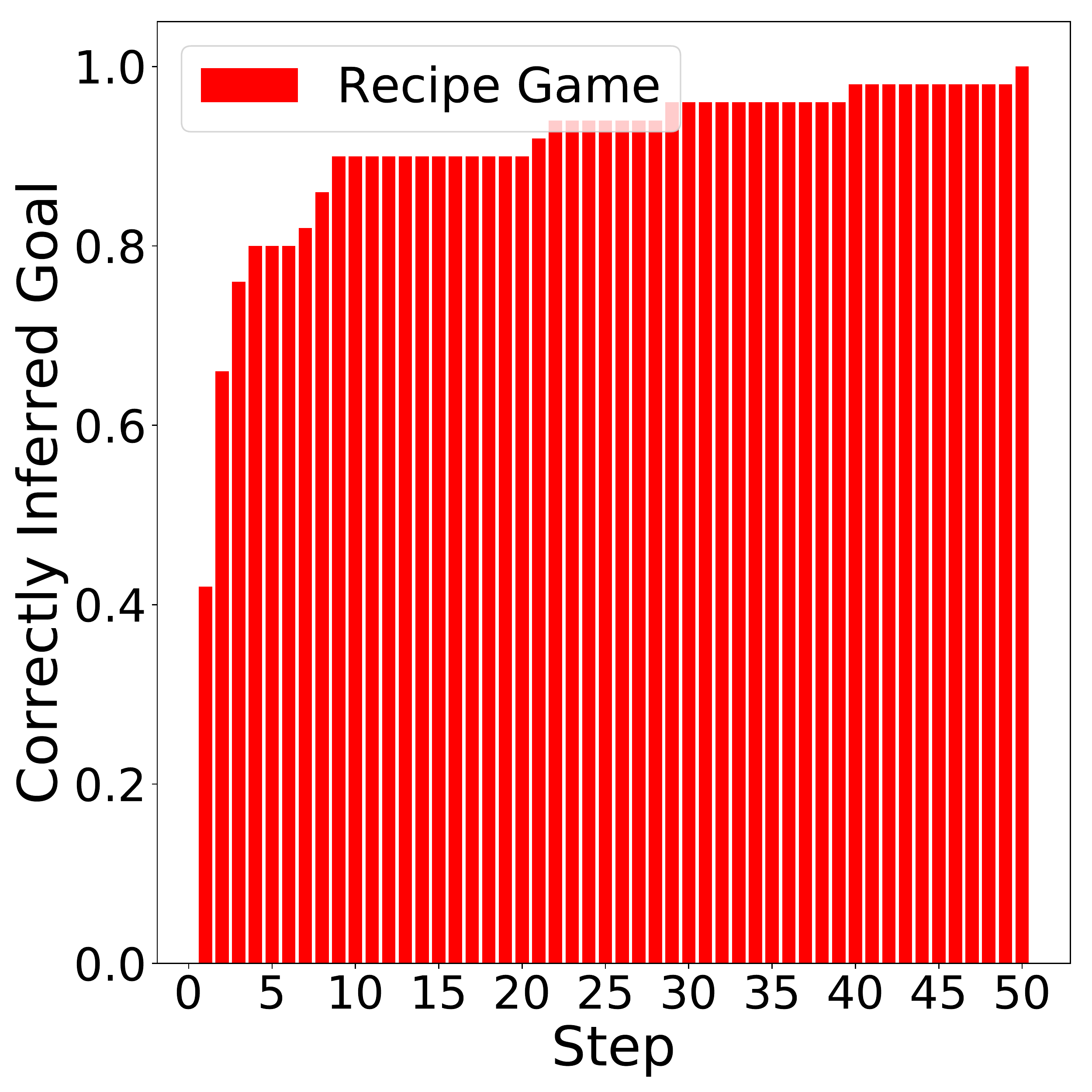}%
		    \includegraphics[width=0.33\linewidth]{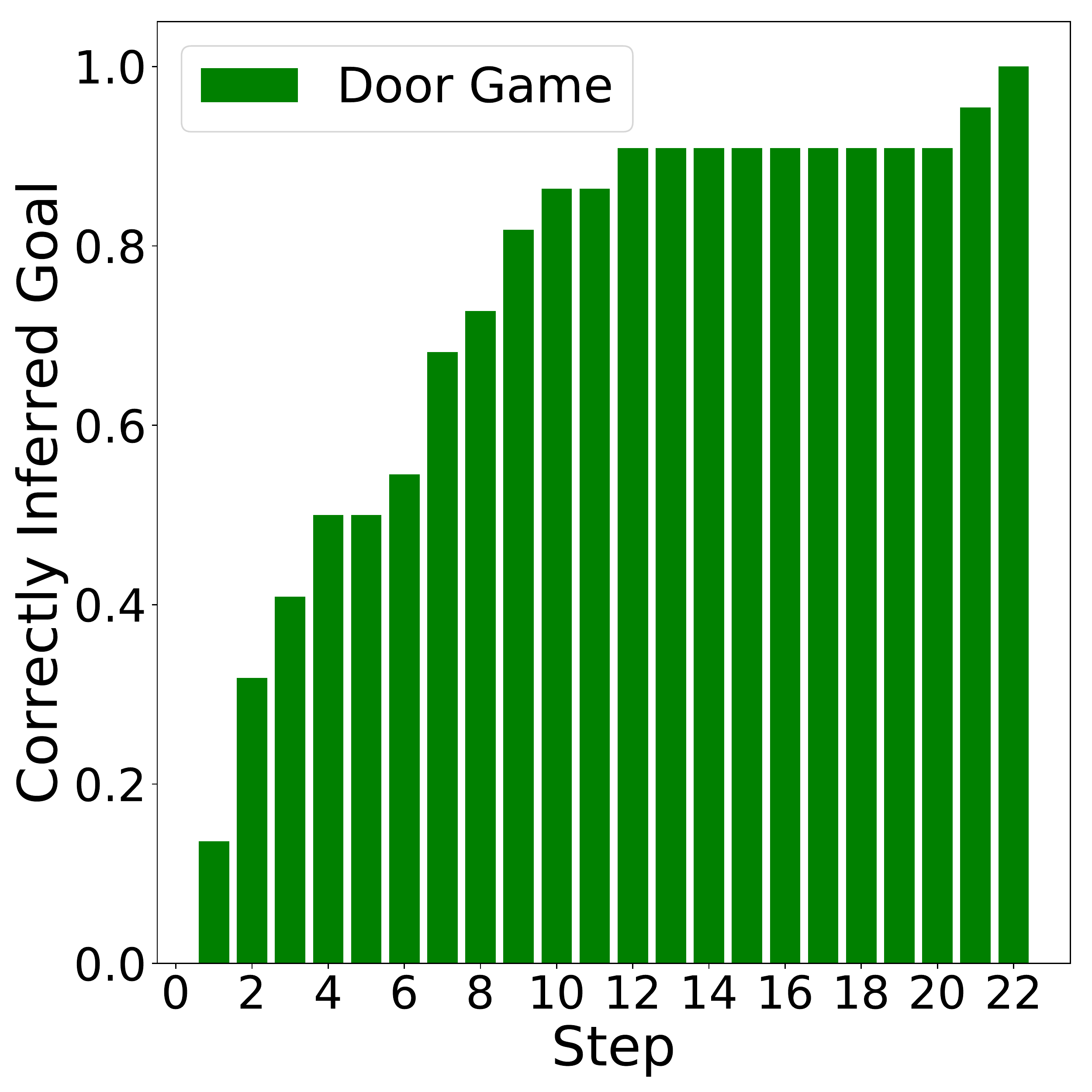}%
		}%
        \caption{\textbf{Inference Step Distribution:} Cumulative distribution of the step $t_{inf}$ at which the goal of the other player is correctly inferred (i.e.
        $\tilde{z}_{other}^t = z_{other}, \forall t \geq t_{inf}$ ) for the Coin (left), Recipe (center) and Door (right) games. We define this step so that $\tilde{z}_{other} = z_{other}$ for all the remaining steps in the game. The distribution is computed over the subset of runs in which the goal is correctly inferred before the end of the game ($\sim70-80\%$ of all runs). A total of 1000 runs with trained SOM models were used to compute this distribution.}  
        % \vspace{-3em}
		\label{fig:inf_step_distr}
	\end{center}
\end{figure}

Figure~\ref{fig:inf_step_distr} shows the cumulative distribution of the step at which the goal of the other player is correctly inferred (and remains the same until the end of the game). The cumulative distribution is computed over the episodes in which the goal is correctly inferred before the end of the game. In the Coin (blue) and Recipe (red) games, $80\%$ of the times the agent correctly infers the goal of the other, it does so in the first five steps. The distribution for the Door (green) game indicates that the agent needs more steps on average to correctly infer the goal. This explains in part why the SOM agent only slightly outperforms the SPP baseline. 
%(which is an approximation of \cite{he2016opponent}'s method). 
If the agent does not infer the other's goal early enough in the episode, it cannot efficiently use it to maximize its reward.

% \begin{figure}[ht]
%     \begin{center}
% 	    \centerline{%
% 		    \includegraphics[width=0.75\linewidth]{figures/percent_inf_step_all.pdf}%
% 		}%
%         \caption{\textbf{Inference Accuracy during an Episode:} Percentage of episodes in which the agent's estimate of the other's goal at that
%         time step is correct for the Coin (blue), Crafting (red), and Door (green) games. ...}  
% 		\label{fig:coin_strategy}
% 	\end{center}
% \end{figure}

% \begin{figure}[ht]
%     \begin{center}
% 	    \centerline{%
% 		    \includegraphics[width=0.75\linewidth]{figures/percent_inf_ep_all.pdf}%
% 		}%
%         \caption{\textbf{Inference Accuracy as a function of Training Epoch:} The mean percentage of episodes in which the agent correctly 
%         infers the goal of the other for the Coin (blue), Crafting (red), and Door (green) games. The estimate of the other's goal is considered correct if the estimate does not change until the end of the game...}  
% 		\label{fig:coin_strategy}
% 	\end{center}
% \end{figure}

\begin{figure}[ht]
    \begin{center}
	    \centerline{%
		    \includegraphics[width=0.33\linewidth]{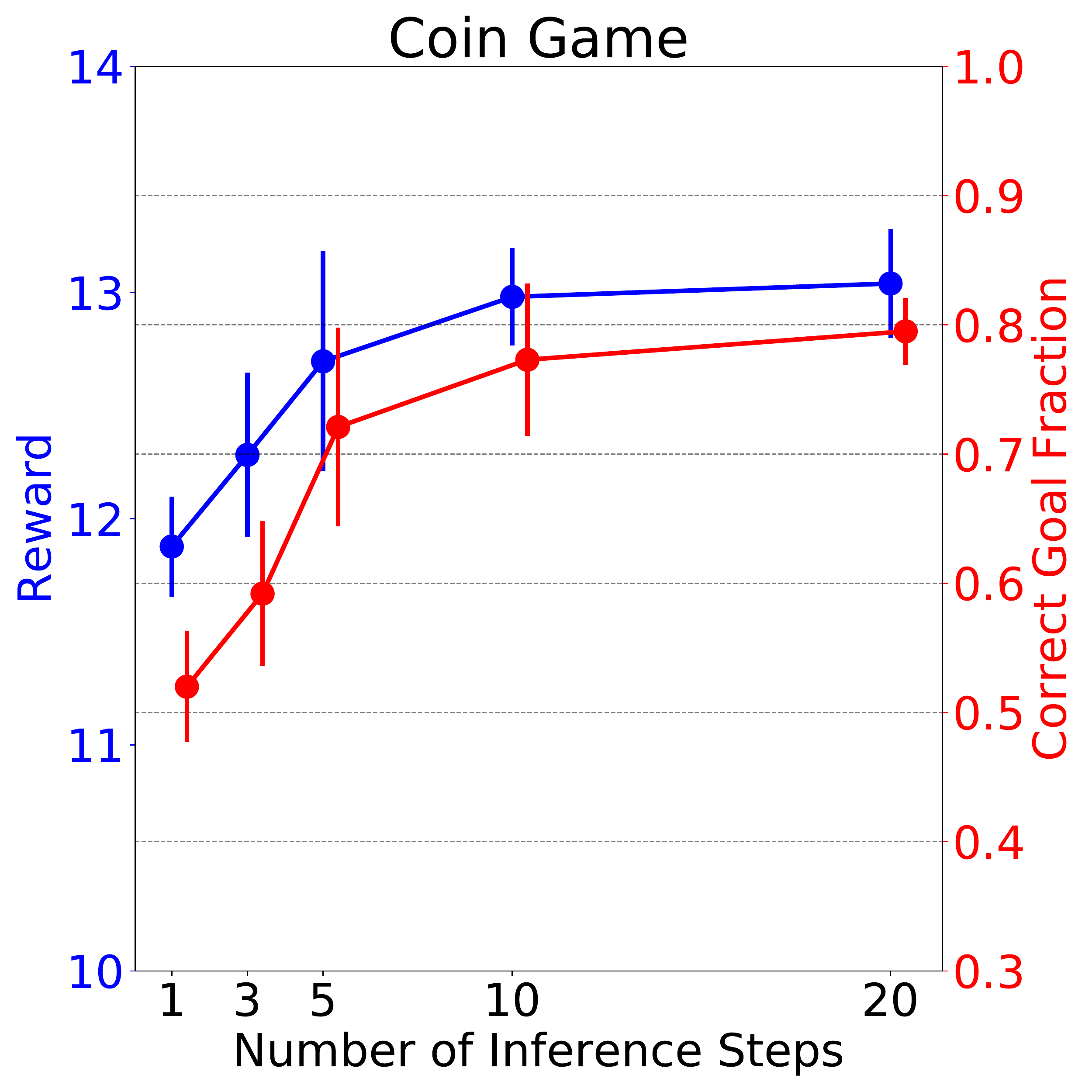}%
		    \includegraphics[width=0.33\linewidth]{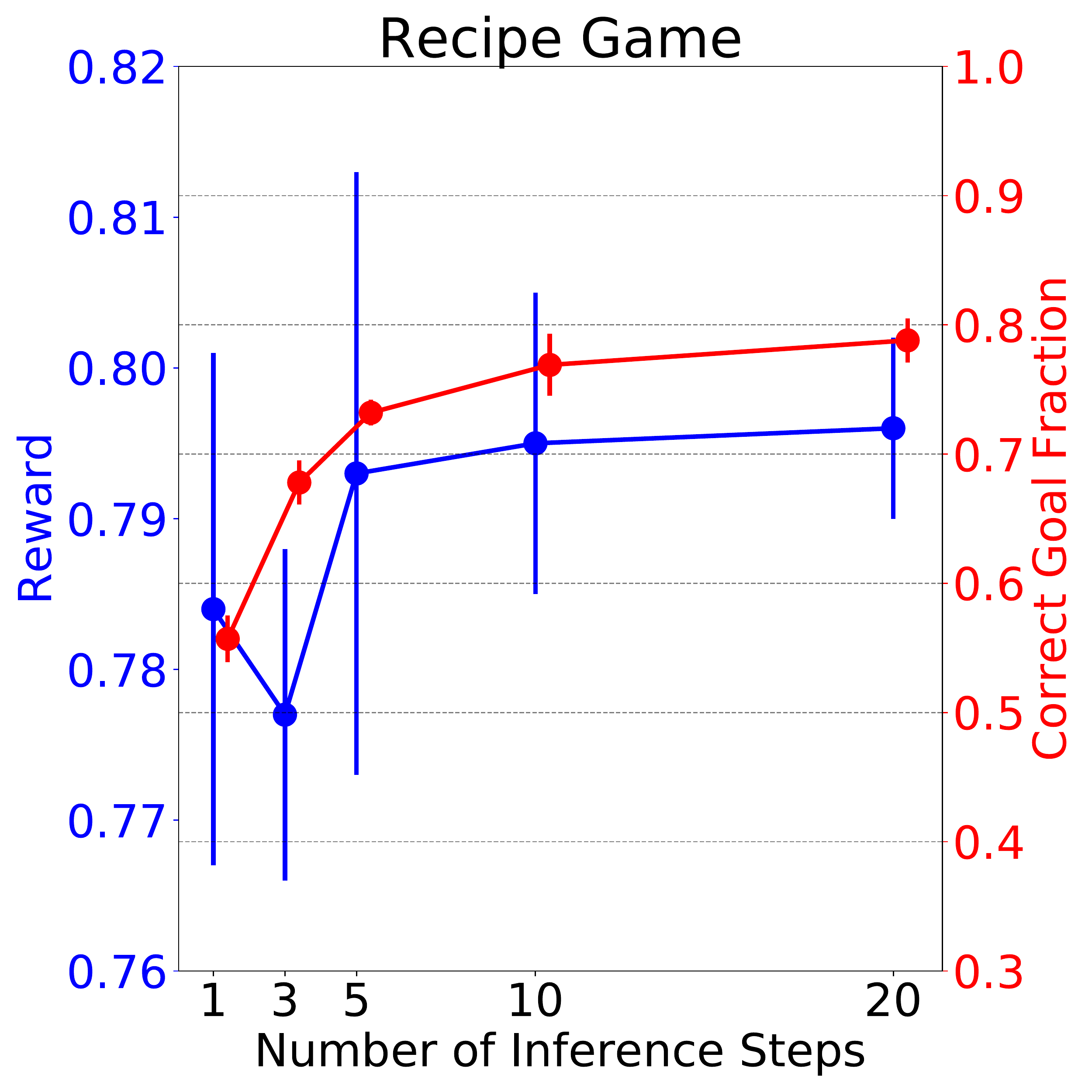}%
		    \includegraphics[width=0.33\linewidth]{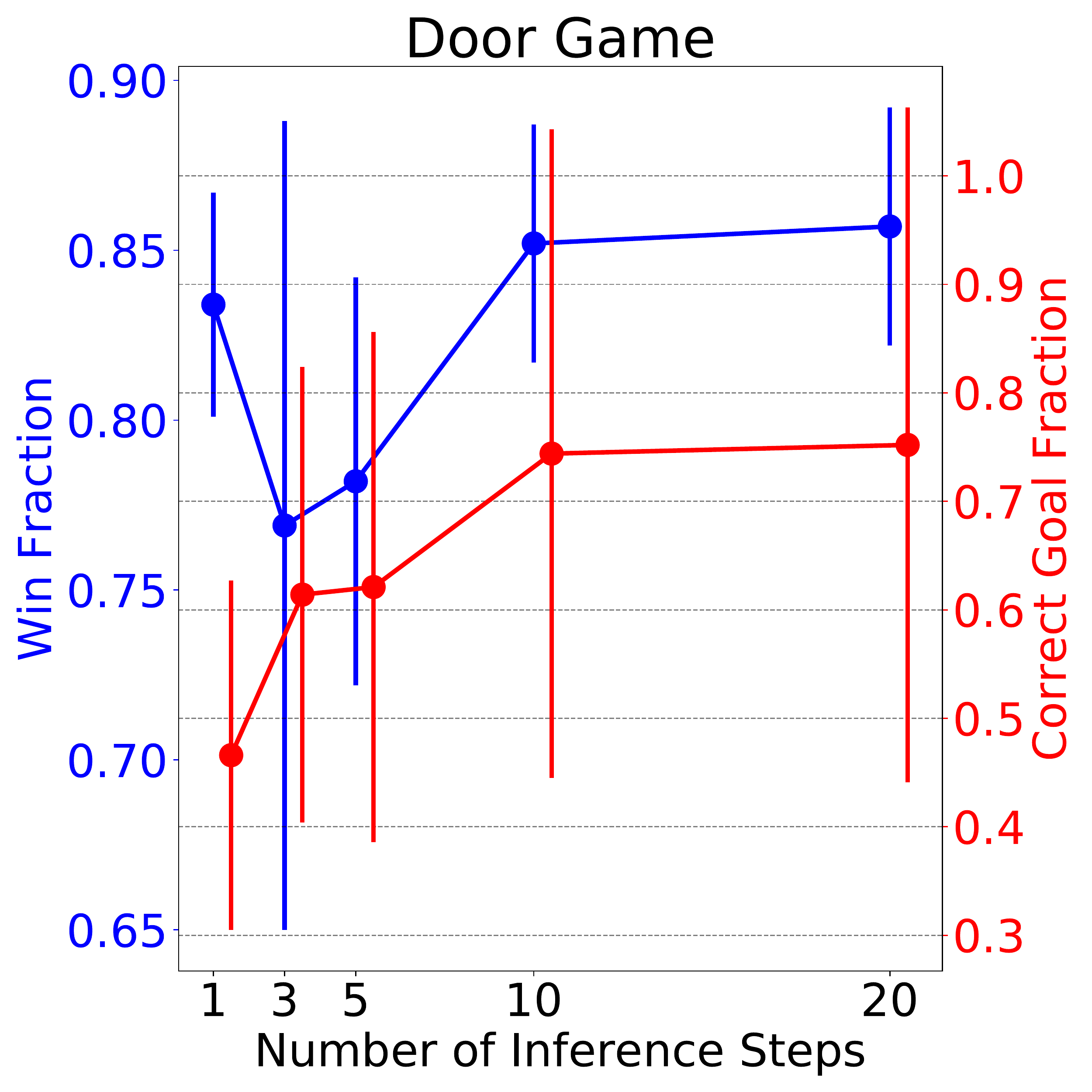}%
		}%
        \caption{\textbf{Performance Variation with Number of Inference Steps:} Average reward (blue) and average
        fraction of episodes in which the goal of the other agent is correctly inferred (red) obtained by the SOM agent as a function of the number of inference steps used for estimating the other's goal for the Coin (left), Recipe (center), and Door (right) games. The error bars represent 1 standard deviation.}
        % \vspace{-3em}
		\label{fig:ninf_var}
	\end{center}
\end{figure}

Figure~\ref{fig:ninf_var} shows how the performance of the agent varies with the number of optimization updates performed on $\tilde{z}_{other}$
at each step in the game. As expected, the agent's reward (blue) generally increases with the number of inference steps, as does the 
fraction of episodes in which the goal is correctly inferred. One should note that increasing the number of inference steps from 10 to 20 
only translates into less than $0.45\%$ performance gain, while increasing it from 1 to 5 translates into a performance gain of 
$6.9\%$ on the Coin game, suggesting that there is a certain threshold above which increasing the number of inference steps will not significantly improve performance. 

\section{Discussion}
\label{discussion}

In this paper, we introduced a new approach for inferring other agents' hidden states 
from their behavior and using those estimates to choose actions. 
We demonstrated that the agents are able to estimate the other players' hidden goals
in both cooperative and competitive settings, which enables them to converge to better policies 
and gain higher rewards. 
In the proposed tasks, using an explicit model of the other player led to better performance 
than simply considering the other agent to be part of the environment. 
% \rob{emphasise generality and simplity of approach: no extra parameters, potentially usable by many different types of policy models etc.}
One limitation of \textsc{SOM} is that it requires a longer training time than the other baselines, 
since we back-propagate through the network at each step. However, their online nature is essential 
in adapting to the behavior of other agents in the environment.

Some of the main advantages of our method are its simplicity and flexibility. This method does not require
any extra parameters to model the other agents in the environment, can be trained with any reinforcement 
learning algorithm, and can be easily integrated with any policy parametrization or network architecture.
The SOM concept can be adapted to settings with more than two players, 
since the agent can use its own policy to model the behavior of any number of agents and infer their goals.
Moreover, it can be easily generalized to many different environments and tasks.

We plan to extend this work by evaluating the models on more complex environments with more than two players, 
mixed strategies, a more diverse set of agent types 
(e.g. agents with different action spaces, reward functions, roles or strategies),
and to model deviations from the assumption that the other player is just like the self. 
% is like the agent.

Other important avenues for future research are to design models that can adapt to non-stationary 
strategies of others in the environment, handle tasks with hierarchical goals, and perform well when playing with
new agents at test time. 

Finally, many research areas could benefit from having a model of other agents that allows reasoning about their
intentions and predicting their behavior. Such models might be useful in human-robot or
teacher-student interactions \cite{dragan2013legibility, fisac2017pragmatic}, 
as well as for value alignment problems \cite{hadfield2016cooperative}. 
Additionally, these methods could be useful for model-based reinforcement learning in multi-agent settings, 
since the accuracy of the forward model strongly depends on the ability of predicting others' behavior. 

\section*{Acknowledgements}
We would like to thank the people who provided feedback along the way:
Ilya Kostrikov, William Whitney, Alexander Rives, 
and Sainbayar Sukhbaatar.

%%%%%%%%%%%%%%%%%%%%%%%%%%% REFERENCES %%%%%%%%%%%%%%%%%%%%%%%%%%%%%%%%%%

\bibliography{references}
\bibliographystyle{icml2018}

\end{document}